%% file: main.tex
\documentclass[conference]{IEEEtran}
\IEEEoverridecommandlockouts
\usepackage{cite}
\usepackage{amsmath,amssymb,amsfonts}

\usepackage[ruled,vlined, linesnumbered]{algorithm2e}

\usepackage{graphicx}
\usepackage{textcomp}
\usepackage{xcolor}
\usepackage{booktabs}
\usepackage{rotating}
\usepackage{multirow}
\usepackage{verbatim}

\usepackage{subfigure}

\usepackage[normalem]{ulem}

\usepackage{bbding}
\usepackage{url}

\usepackage{amsthm} 
\newtheorem{theorem}{Theorem}

\usepackage{bbding}

\usepackage{hyperref}

\usepackage{balance}

\newcommand{\oursfull}{\underline{S}treaming Edge \underline{P}artitioning and Parall\underline{e}l Acc\underline{e}leration for Temporal Interaction Graph Embe\underline{d}ding (SPEED)}

\newcommand{\oursfullp}{\underline{S}treaming \underline{E}dge \underline{P}artitioning Component (SEP)}

\newcommand{\oursfulld}{\underline{P}arallel \underline{A}cceleration \underline{C}omponent (PAC)}

\newcommand{\oursfullnolinep}{Streaming Edge Partitioning Component}

\newcommand{\oursfullnolined}{Parallel Acceleration Component}

\newcommand{\ours}{SPEED}
\newcommand{\oursp}{SEP}
\newcommand{\oursd}{PAC}

\def\BibTeX{{\rm B\kern-.05em{\sc i\kern-.025em b}\kern-.08em
    T\kern-.1667em\lower.7ex\hbox{E}\kern-.125emX}}
\begin{document}

\title{

\mbox{SPEED: Streaming Partition and Parallel Acceleration}
\mbox{for Temporal Interaction Graph Embedding}

}

\author{\IEEEauthorblockN{Xi Chen$^\dagger{}^\ast$, Yongxiang Liao$^\dagger{}^\ast$, Yun Xiong$^\dagger{}^\S$, Yao Zhang$^\dagger$, Siwei Zhang$^\dagger{}$, Jiawei Zhang$^\ddagger{}$, Yiheng Sun$^\|$}
\IEEEauthorblockA{\textit{$^\dagger$Shanghai Key Laboratory of Data Science, School of Computer Science, Fudan University, Shanghai, China} \\
\textit{$^\ddagger$IFM Lab, University of California, Davis, CA, United States}\\
\textit{$^\|$Tencent Weixin Group, Shenzheng, China}\\
\{x\_chen21, liaoyx21\}@m.fudan.edu.cn, \{yunx, yaozhang\}@fudan.edu.cn,\\ swzhang22@m.fudan.edu.cn, jiawei@ifmlab.org, sunyihengcn@gmail.com
}
}

\maketitle

\renewcommand{\thefootnote}{\fnsymbol{footnote}}
\footnotetext[1]{Xi Chen and  Yongxiang Liao contributed equally to this paper.} 
\footnotetext[4]{Yun Xiong is the corresponding author.} 

\begin{abstract}
Temporal Interaction Graphs (TIGs) are widely employed to model intricate real-world systems such as financial systems and social networks.
To capture the dynamism and interdependencies of nodes, existing TIG embedding models need to process edges sequentially and chronologically.
However, this requirement prevents it from being processed in parallel and struggle to accommodate burgeoning data volumes to GPU. Consequently, many large-scale temporal interaction graphs are confined to CPU processing.
Furthermore, a generalized GPU scaling and acceleration approach remains unavailable. To facilitate large-scale TIGs' implementation on GPUs for acceleration, we introduce a novel training approach namely \oursfull. The \ours~is comprised of a \oursfullnolinep~(\oursp) which addresses space overhead issue by assigning fewer nodes to each GPU, and a \oursfullnolined~(\oursd) which enables simultaneous training of different sub-graphs, addressing time overhead issue.
Our method can achieve a good balance in computing resources, computing time, and downstream task performance. Empirical validation across 7 real-world datasets demonstrates the potential to expedite training speeds by a factor of up to 19.29x. Simultaneously, resource consumption of a single-GPU can be diminished by up to 69\%, thus enabling the multiple GPU-based training and acceleration encompassing millions of nodes and billions of edges. Furthermore, our approach also maintains its competitiveness in downstream tasks. 

\end{abstract}


\section{Introduction}
\label{sec:intro}
\begin{figure}[!t]
\centering
\includegraphics[width=1\linewidth]{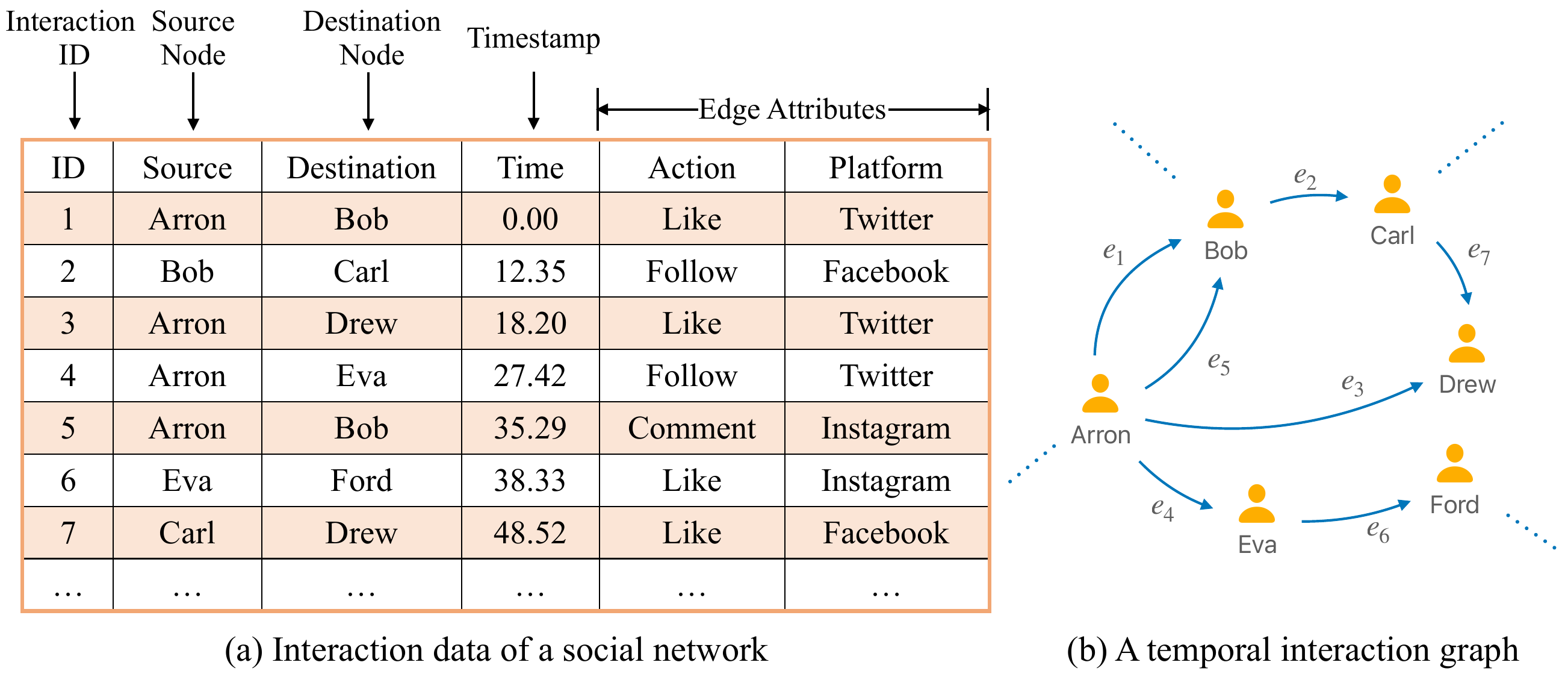}
\caption{Interaction data of a social network and its corresponding temporal interaction graphs (TIG). $e_i$ in (b) refers to the edge which contains the information of time and edge attributes.}
\label{fig:tig_tab}
\end{figure}

Real-world systems featuring sequences of interaction behavior with timestamps—such as social networks, financial trades, and recommendation systems—can all be conceptualized as Temporal Interaction Graphs (TIGs) \cite{jodie, dyrep, tgat, tgn, tiger}. Given a series of timestamped interactions, existing TIG embedding models \cite{jodie, dyrep, tgat, tgn, tiger} represent the objects as nodes and the interaction behaviors among objects with time information as edges, an example is demonstrated in Fig.\ref{fig:tig_tab}. These models encode historical interaction, i.e., event, information in nodes message and memory modules \cite{jodie, dyrep, tgat, tgn, tiger}. The nodes' embedding can be generated and applied to various downstream tasks by consuming nodes' history information.

As these real-world systems evolve, the associated data scale will also expand correspondingly. Previous research on temporal interaction graph embedding \cite{jodie, dyrep, tgat, tgn, tiger} has primarily focused on enhancing downstream task performance on small datasets, often neglecting the training of large-scale data and efficiency considerations. Based on the updating mechanism of the memory module of existing models, as the number of nodes increases, a larger amount of memory information will need to be stored, leading to higher computing resource requirements. Simultaneously, an upsurge in interaction behaviors leads to considerable overhead in both computational memory resources and time costs.

\begin{figure*}[htbp]
	\centering
	\subfigbottomskip=2pt
	\subfigcapskip=-5pt
	\subfigure[Traditional training on Single-GPU \cite{tgat, tgn, tiger}]{
		\includegraphics[width=0.95\linewidth]{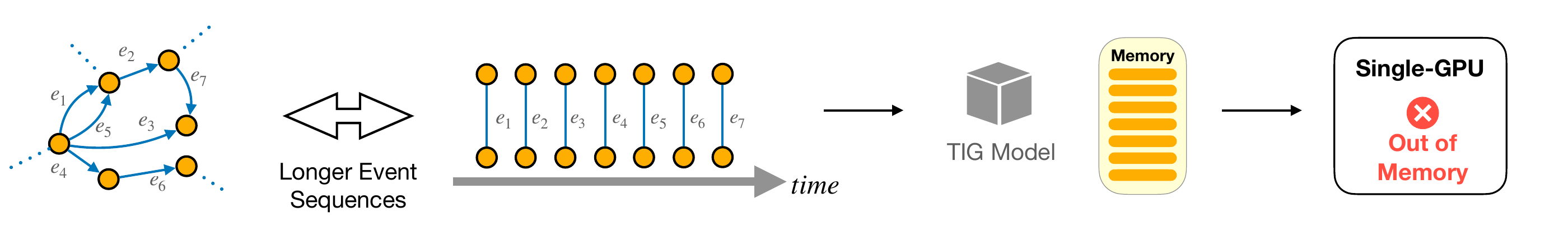}}
	\subfigure[Training with Partitioned Sub-Graphs on Multi-GPUs with Shared Nodes \textbf{[This Paper]}]{
		\includegraphics[width=0.95\linewidth]{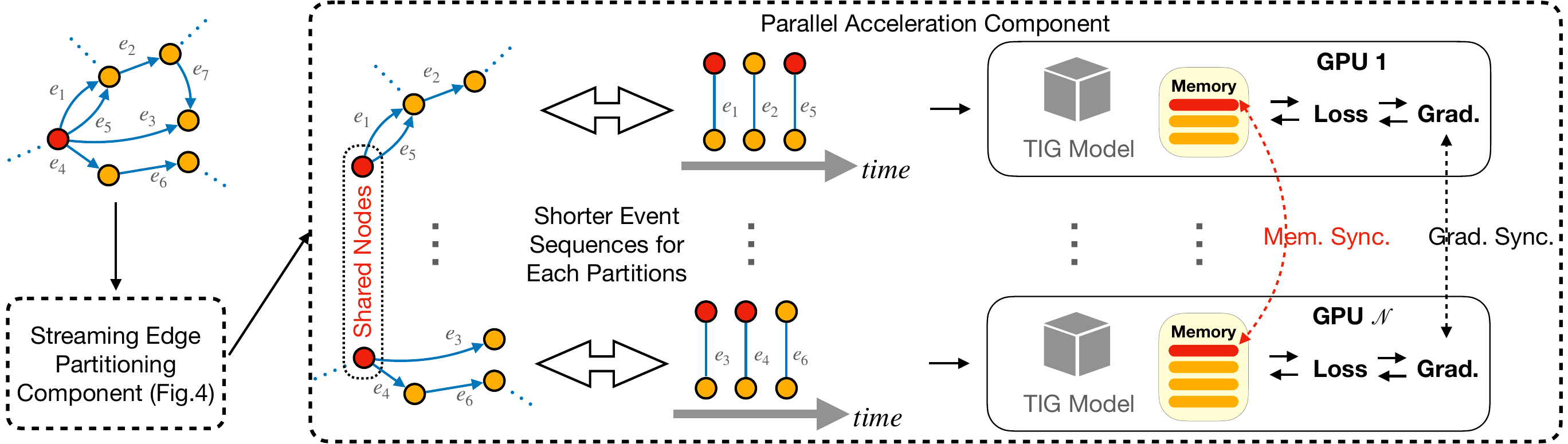}}
	\caption{A comparison between our training approach and conventional single-GPU training. Our approach has much lower training memory and time costs by deploying fewer edge data on each GPU and executing the training in parallel. Benefiting from \oursfullnolinep~(\oursp) (illustrated in Fig.\ref{fig:sgp}), we can also accommodate to larger datasets, since the memory required for nodes memory module in our approach per GPU is smaller than in traditional single-GPU training. Note that the outputs of SEP are the partitioned sub-graphs which are the inputs of \oursfullnolined.}
\label{fig:flow}
\end{figure*}

Traditional single-GPU or CPU training methods \cite{jodie, dyrep, tgat, tgn, tiger} will encounter significant challenges when handling large-scale temporal interaction graphs due to substantial time and space overhead. The time overhead is considerable, especially when dealing with graphs characterized by a large number of edges. From a spatial perspective, a single-GPU will struggle to accommodate to and store memory information for an extensive number of nodes. 
Consequently, we aspire to construct a training approach capable of addressing both time and space overhead problems.

A straightforward solution to mitigate time overhead is through parallel acceleration using multiple GPUs.
However, conventional TIG models \cite{jodie, dyrep, tgat, tgn, tiger} pose significant challenges for parallel training of TIGs. Unlike in static graphs, the message-passing operations in TIGs must adhere to time-based restrictions, preventing a node from gathering information from future neighbors.
This leads to \textbf{\textit{Challenge 1}}: Temporal and structural dependencies\textemdash{}how to overcome time-based message-passing restrictions in TIGs' implementation and train TIG models in parallel while preserving the complex interplay between temporal and structural dependencies. 
Additionally, in TIGs, all events of nodes are interconnected \cite{edge, tiger}. This necessitates the model to traverse past edges sequentially and chronologically to keep nodes memory up-to-date. It will result in \textbf{\textit{Challenge 2}}: Training interaction sequences in parallel\textemdash{}how to handle multiple temporal interaction sequences with interconnected events while preventing information loss among them and updating nodes' memory in a distributed parallel training manner. 
Existing TIG models are struggle to handle the large-scaled TIG data studied in this paper, due to the bottleneck of training in parallel \cite{edge}.


The space overhead issue primarily originates from the memory module. In most existing baseline methods, a memory slot is maintained for each node to update its representation. While storing memory module in GPU can accelerate computation, as the number of nodes increases, the storage for these nodes' memory may also lead to excessive GPU memory consumption. This poses \textbf{\textit{Challenge 3}}: Space overhead caused by memory module\textemdash{}how to accommodate and manage nodes' memory storage for large-scale TIGs on GPUs with a limited memory size.

In response to all above challenges, we introduce an effective and efficient approach, namely \oursfull. The \ours~consists of two functional components, i.e., the \oursfullp~and the \oursfulld.
The TIG partitioning strategy entails assigning fewer graph nodes to each GPU, thereby regulating the nodes memory module's GPU memory consumption. This helps to address the issue of space overhead. Another advantage of this lies in the reduction of corresponding edges per GPU. Utilizing the multi-GPU parallel component enables simultaneous training of different edge sequence data, thereby accelerating the training process. 


The graph partitioning is crucial for handling large-scale TIGs. 
However, as shown in Tab.\ref{tab:1}, current graph partitioning algorithms fail to simultaneously satisfy the following requirements: i) temporal information consideration: they neglect the temporal aspect of TIG, disregarding the diverse messages brought by edges at varying timestamps; ii) low replication factor: replicated nodes are added to decrease information loss, however, these algorithms lack control over the number of these nodes, leading to space overhead issues; iii) load balancing: ensuring an even distribution of edges and nodes to balance the training time and resource usage across different GPUs; iv) good scalability: a requirement for low algorithm overhead and the ability to scale efficiently to large-scale temporal interaction graphs.

\input{tab_1}

Hence, we propose the \oursfullp.
Firstly, we introduce an exponential time decay strategy to incorporate temporal information. It aims to capture the recent trend of interactions between nodes. Then, we estimate the centrality of each node by aggregating the weights of its historically connected edges.
Moreover, to minimize the replication factor, we designate nodes with the top-$k$ centrality values as $hubs$. We treat the input graph as stream of edges, sequentially assigning edge to a specific partition. During this phase, only nodes classified as $hubs$ can be duplicated across different partitions as ``shared nodes''. It ensures that the vital information will be maintained across all partitions without unnecessary replication of data.
Furthermore, we apply a greedy heuristic to maintain the load balance among partitions.
The combination of the above strategies enables SEP to satisfy all the requirements previously outlined in Tab.\ref{tab:1}. 

Given the time-sensitive nature of TIGs, the training data is fed into the model in a chronological order, aligned with the timestamp of each edge. Therefore, if the graph is merely divided and allocated to different GPUs for independent computation, the GPU processing for edges with later timestamps will need to wait for those with earlier timestamps. This creates a bottleneck for the multi-GPU training for TIGs.

To mitigate GPU waiting and adjust the \oursp~component, we propose the \oursfulld. Initially, each sub-graph is assigned to a distinct GPU, ignoring inter-node edges across different GPUs. This means some edges get deleted, causing potential information loss and model effectiveness reduction.
To counterbalance this, we take advantage of the shared nodes.
The memory of shared nodes is synchronized after each training epoch as a compensatory measure. We further introduce a random shuffling approach.
Specifically, we partition the graph into smaller sub-graphs and then shuffle and amalgamate them to fit the number of GPUs.
As this precedes every epoch, various ``deleted'' edges between small partitions can be recovered by merging them into larger partitions and trained across epochs. The \oursd~allows us to train different sub-graphs in parallel on multi-GPUs, and alleviate information loss.

\begin{figure}[!tbp]
\includegraphics[width=1\linewidth]{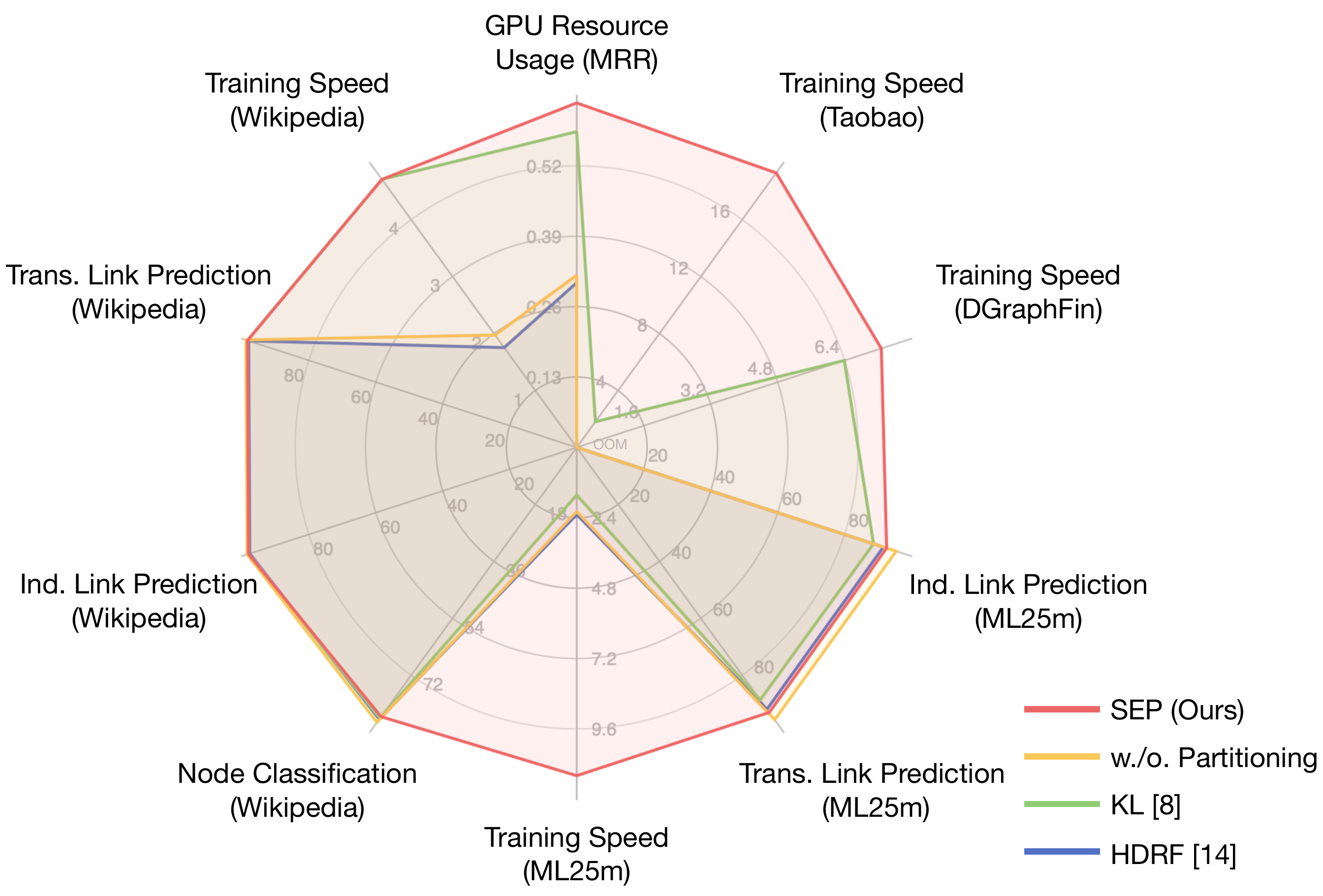}
\caption{Comparison between graph partitioning methods. The axes are arranged counterclockwise, with graph sizes increasing. Our method outperforms others in terms of training speed (especially as the graph size increases) and GPU resource utilization (on one GPU), while achieving comparable performance on downstream tasks (link prediction in both transductive and inductive styles, as well as node classification on representative datasets. For full experimental results, please refer to Sec.\ref{sec:exps}). The data are based on averaged experimental results with the TIGE \cite{tiger}. MRR refers to mean reciprocal ranking; Training Speed refers to training speed-up(x) compared with CPU training; OOM refers to out-of-memory. Since the maximum number for each axis differs, the ticks may also vary between axes.}
\label{fig:radar}
\end{figure}

Our \ours~overcomes the bottlenecks of the existing TIG embedding models for large-scale graphs in terms of computing efficiency (time) and computing resources (device memory). To further provide readers with the outline of the \ours, Fig.\ref{fig:flow} offers a comparative illustration of our training approach and the conventional single-GPU approach, highlighting our approach's enhanced capacity to handle large datasets. Additionally, as demonstrated in Fig.\ref{fig:radar} and supported by extensive experimental studies, our approach significantly accelerates the training process while maintaining competitive performance on downstream tasks.

The contributions of this paper are as follows:
\begin{enumerate}
\item{We propose a novel large-scale Temporal Interaction Graph embedding approach which achieves a balance among computational resources, time costs and downstream task performance.
%
}
\item{We design a steaming partitioning strategy, specifically tailored for parallel training. This strategy not only captures time information, but also maintains load balance and low replication factor.}
\item{We present a parallelable acceleration method for training graphs with billions of temporal interactions using multi-GPUs.
With the partitions shuffling and memory synchronizing across sub-graphs, our approach alleviates the information loss raised by graph partitioning.}
\item{
Extensive experimental results demonstrate that the proposed approach significantly expedites training speeds and reduces resource consumption while maintaining its competitiveness in downstream tasks.
}
\end{enumerate}

\section{Proposed methods}
\label{sec: methods}

\begin{figure*}[!t]
\centering
\includegraphics[width=1\linewidth]{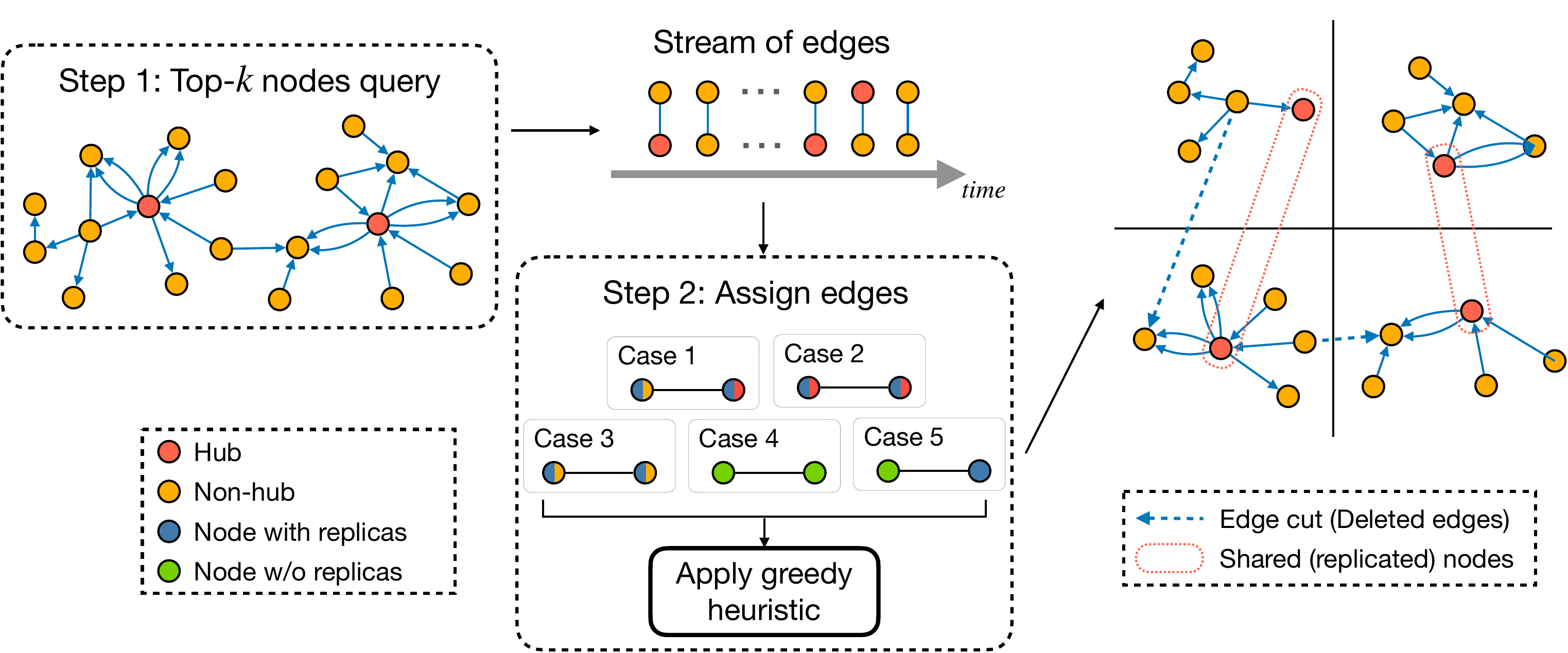}
\caption{Overview of the workflow of \oursfullnolinep. Please refer to Alg.\ref{alg:partitioning} for details of the different cases in step 2.}
\label{fig:sgp}
\end{figure*}

\subsection{Notations}
Given a set of nodes $\mathcal{V} = \{1,2, \dots, N \}$, nodes features are noted as $\mathbf{v}_i\in \Re^d$. $\mathcal{E}=\{e_{ij}(t)=(i, j, t) | i,j \in \mathcal{V} \}$ is a series of interaction behaviours (as shown in Fig.\ref{fig:tig_tab}), where interaction event $e_{ij}(t)=(i, j, t)$ happens between nodes $i,j$ at time $t\in [0,t_{\max}]$. The interaction feature is noted as $\mathbf{e}_{ij}(t) \in \Re^d$. Then, we have a temporal interaction graph $\mathcal{G}=(\mathcal{V}, \mathcal{E})$, which stores the timestamps in edge features. $\mathcal{E}_t = \{(i,j,\tau)\in \mathcal{E} | \tau < t \}$ contains interaction events record before time $t$. In the case of a non-attributed graph, we make the assumption that node and edge features are zero vectors. For the sake of simplicity, we maintain a consistent dimensionality of $d$ for node and edge features, as well as for node representations. 

\subsection{\oursfullp}
\label{sec:TIG part}
As graph partitioning is a preprocessing step in distributed training, the quality of the partitioning may have great impact on the quality of the distributed training.
To attain the objectives of the four dimensions as outlined in Tab. \ref{tab:1}, we employ the node-cut based streaming partitioning algorithm, complemented by two key innovations: First, in order to introduce temporal information, we employ an \emph{exponential time decay} strategy to define the centrality of nodes. Second, we control the number of shared nodes in the process of edge assignment to avoid high replication factor while minimizing the information loss. An illustration can be found in Fig.\ref{fig:sgp}.

\textbf{Calculation of node's centrality.} In TIG model training, whenever an event occurs (i.e., an edge is added), the information from that edge serves as an input, updating the representations of the involved nodes. Temporal neighbor sampling is essential to ensure that a node can acquire ample information about its neighbors, thus mitigating the ``staleness problem'' \cite{tgn}. A common method for temporal neighbor sampling is sampling only the most recent neighbors. Our intuition is that the more recent events often have a greater impact on node's future actions \cite{tgat}. 
Therefore, before the graph partitioning phase, we will assign a greater weight to more recently appearing edges. Based on the observation above, we propose an \emph{exponential time decay} strategy for edge weight computation, which is commonly applied in temporal neighbor sampling \cite{pint,cawn,zebra}. Concurrently, the node's centrality is determined by aggregating the weights of all its historically connected edges.
Let $\mathcal{T}(i)$ denotes the set of timestamps for all historical edges of node $i$ and $t_{max}$ as the last timestamp. For node $i$, its centrality is defined as:
\begin{equation}
    Cent(i) = \sum_{t \in \mathcal{T}(i)} exp(\beta(t-t_{max})),
\end{equation}
where $\beta \in (0, 1)$ is a scalar hyper-parameter. 

\begin{figure}[!t]
\centering
\includegraphics[width=1\linewidth]{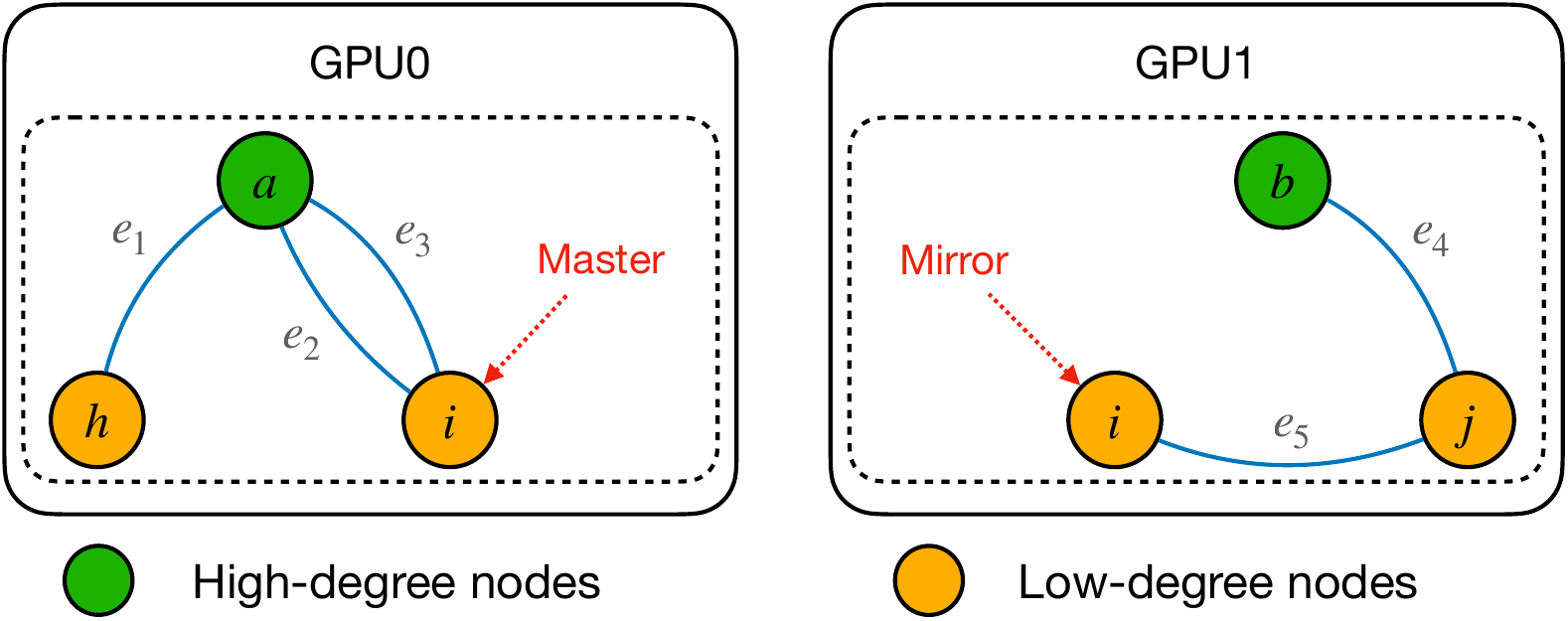}
\caption{An example of a low-degree node $i$ being replicated in HDRF\cite{hdrf}. The order of occurrence of the edges is $(e_1, e_2, e_3, e_4, e_5)$. The mirror node $i$ is replicated in GPU1 when edge $e_5 = (i,j)$ appears because HDRF tends to replicate node with higher degree.}
\label{fig:exa}
\end{figure}

\input{alg_partitioning}

\textbf{Streaming node-cut graph partitioning algorithm.} The standard greedy heuristic introduced in \cite{greedy} may yield highly imbalanced partition sizes and a high replication factor \cite{hdrf}, as it treats nodes with varying degrees as equals. In the HDRF algorithm \cite{hdrf}, the degree of nodes is factored into the greedy heuristic. But the HDRF only considers the static graph and takes it as stream of edges that are input in some order like DFS or BFS. In real-world TIG partitioning, however, the appearance of some unpredictable edges can cause a large number of low-degree nodes to be replicated. As shown in Fig.\ref{fig:exa}, low-degree node $i$ is replicated in GPU1 when edge $e_5 = (i,j)$ appears. 

To avoid such situations, we first query the top-$k$ most important nodes (a hyper-parameter of our method, noted as $top_k$) as \emph{hubs}, based on the node centrality calculation in the previous step. Then we take the input TIG as stream of edges and sequentially assign edge to a specific partition. During the edge assigning phase, we restrict replication to nodes in $hubs$, thereby reducing the replication factor and preserving edge information as much as possible. Moreover, to maintain load balance, we employ a greedy heuristic
at lines 8 and 16 in Alg.\ref{alg:partitioning}. Specifically, when an input edge $e = (i,j,t)$ is being processed, the normalized centrality value of nodes $i$, $j$ are defined as:
\begin{equation}
    \theta(i) = \frac{Cent(i)}{Cent(i)+Cent(j)} = 1 - \theta(j).
\end{equation}
Then we compute a score $C(i,j,p)$ for each partition $p$ and greedily assign the edge to the partition with the maximum score $C$ defined as follows:
\begin{equation}
    C(i,j,p) = C_{REP}(i,j,p) + C_{BAL}(p).
\label{eq:3}
\end{equation}
The first term $C_{REP}$ is to ensure that edge is assigned to the partition where the lower centrality node asides (line 8 in Alg.\ref{alg:partitioning}) or to the partition where the node already be assigned (line 16 in Alg.\ref{alg:partitioning}). The second term $C_{BAL}$ is to maintain load balancing by assigning edge to the smallest partition.
More specifically, $C_{REP}$ is defined as:
\begin{equation}
    C_{REP}(i,j,p) = h(i,p) + h(j,p).
\end{equation}
For node $i$ and partition $p$, $h(i,p)$ is defined as:
\begin{equation}
    h(i,p) = 
    \begin{cases}
    1 + (1-\theta(i)), & if \; p \in A(i)\\
    0, & otherwise
    \end{cases},
\end{equation}
where $A(i)$ denotes to the set of partitions to which node $i$ has been assigned.
The second term $C_{BAL}$ is defined as:
\begin{equation}
    C_{BAL}(p) = \lambda \cdot \frac{maxsize - |p|}{\epsilon + maxsize -minsize}.
\end{equation}
The parameter $\lambda$ ($\lambda > 0$) manages the impact of load balancing in greedy heuristic. Meanwhile, $\epsilon$ is a small constant added to avoid zero denominators in the calculations, with $maxsize$ and $minsize$ defining the upper and lower limits of the partition size, respectively.\par
At lines 17-22 in Alg.\ref{alg:partitioning}, after edge assigning, we add the nodes with replicas in more than one partition to the shared nodes list, which is used as input for subsequent distributed training.

\textbf{Theoretical Analysis.} 
In this section, we perform a theoretical analysis of \oursp, particularly examining the worst-case scenarios for the replication factor ($RF$) and edge cuts ($EC$).

\textbf{Metrics.} We use the metrics as follows:
\begin{equation}
    RF = \frac{Total\;node\;replicas}{Total\;number\;of\;nodes},
\end{equation}
\begin{equation}
    EC = \frac{Total\;edge\;cuts\;between\;partitions}{Total\;number\;of\;edges}.
\end{equation}
\begin{theorem}
When partitioning a graph with $|\mathcal{V}|$ nodes on $|\mathcal{P}|$ partitions, our algorithm achieves a upper bound of $RF$ as:
\begin{equation}
    RF \le k|\mathcal{P}| + (1-k).
\end{equation}
\end{theorem}
\begin{proof}
The first term signifies the replicas generated by the fraction $k$ of nodes (referred as $hubs$) with the highest centrality in the graph. In the worst-case scenario, all nodes within the $hubs$ have duplicated copies across all partitions. The second term consider the replicas created by non-$hubs$. In our algorithm, they each can create at most one replica.
\end{proof}
Cohen et al. \cite{attack} demonstrated that when a fraction $k$ of nodes with the highest degrees is removed from a power-law graph (along with their edges), the maximum node degree $M$ in the remaining graph can be approximated as:
\begin{equation}
    M = mk^{\frac{1}{1-\alpha}},
\end{equation}
where $m$ is the minimum node degree in the graph and $\alpha$ is a parameter that indicates the skewness of the graph.

To provide a theoretical upper bound on $EC$, we directly employ the degree of a node as its centrality value.
\begin{theorem}
    When partitioning a power-law graph with $|\mathcal{V}|$ nodes and 
 $|\mathcal{E}|$ edges on $|\mathcal{P}|$ partitions, our algorithm achieves a upper bound of $EC$ as:
 \begin{equation}
     EC \le \frac{1}{|\mathcal{E}|}\sum_{q=0}^{|\mathcal{V}|(1-k)-1} m(k+\frac{q}{|\mathcal{V}|})^{\frac{1}{1-\alpha}}.
 \end{equation}
\end{theorem}
\begin{proof}
    According to Alg.\ref{alg:partitioning}, edge dropping only occurs during the execution of $Case\;3$. It happens when both nodes are non-$hubs$ and have replicas in different partitions. Therefore, edges connected to $hubs$ are preserved. The worst-case scenario is that all edges connecting two non-$hubs$ are dropped. This scenario occurs when all non-$hubs$ are connected to $hubs$ upon their first appearance, and all edges between two non-$hubs$ cross partitions. Given that the largest degree of the graph excluding $hubs$ is $mk^{\frac{1}{1-\alpha}}$, when we remove a non-$hub$ node with the highest degree, the highest degree in the remaining graph is $m(k+\frac{1}{|\mathcal{V}|})^{\frac{1}{1-\alpha}}$. Therefore, we can bound $EC$ by counting all edges connecting two non-$hubs$.
\end{proof}

\subsection{\oursfulld}

\begin{figure}[!t]
\centering
\includegraphics[width=0.8\linewidth]{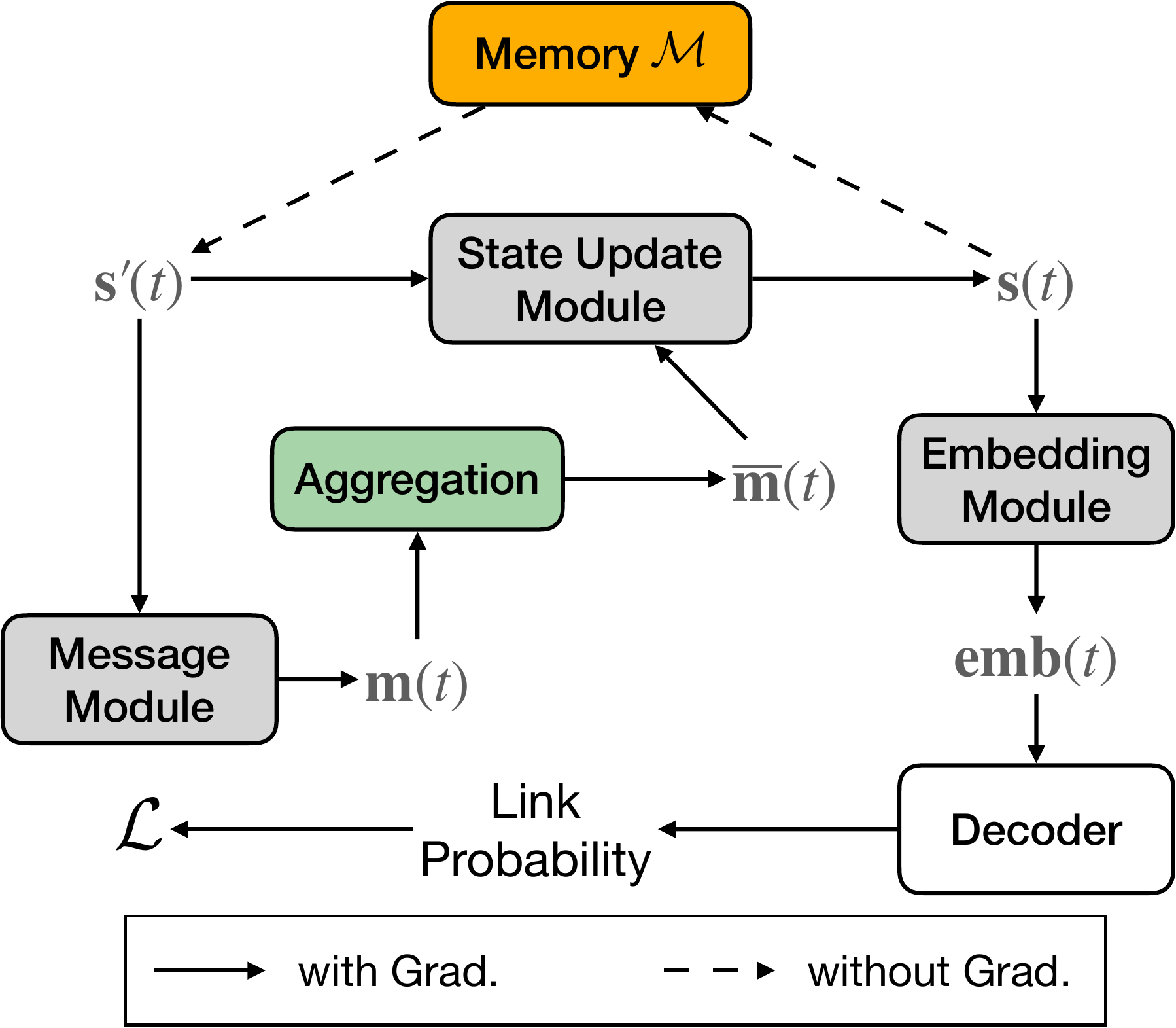}
\caption{Illustration of data flows of TIG models \cite{tiger}. Note that the memory module is constantly being updated.}
\label{fig:tig}
\end{figure}

\textbf{Training approach for TIGs.}
The data flows of most models are shown in Fig.\ref{fig:tig}.
A TIG model typically adopts an Encoder-Decoder structure, with the Encoder composed of four key modules.
To avoid repeat calculation, the \textit{Memory Module} $\mathcal{M} \in \Re^{N \times d}$ is employed to capture the historical interaction information for each node $i$, represented as $\mathcal{M}_i$. 
The \textit{Message Module} is used to compute the current state, i.e., message $\mathbf{m}_i(t)$ of nodes. For an interaction event $e_{ij}(t)$, messages are computed by the previous states $\mathbf{s}'_i(t)$, $\mathbf{s}'_j(t)$, event feature vector $\mathbf{e}_{ij}(t)$, and time encoding computed by $\Delta t$. The message computing functions ($MSG$) are learnable and can be choose from different neural networks or just simply concatenation the inputs. We use node $i$ as an example, where $\Phi$ is the time encoding function \cite{tgat}:
$\mathbf{m}_i(t) = MSG(\mathbf{s}'_i(t), \mathbf{s}'_j(t), \Phi(\Delta t), \mathbf{e}_{ij}(t))$.
Previous states are fetched from node memory, i.e., $\mathbf{s}'_i(t)\leftarrow\mathcal{M}_i$ and $\mathbf{s}'_j(t)\leftarrow\mathcal{M}_j$. All messages in the same batch of a node can be aggregated by an aggregation function, which can be simply mean message or other neural networks, e.g., RNN, and output the aggregated message $\overline{\mathbf{m}}_i(t)$.
After an event which involve node $i$ happened, the new node representation can be updated by the node state before the event $\mathbf{s}'_i(t)$ and the aggregated message $\overline{\mathbf{m}}_i(t)$ by applying the \textit{State Update Module}:
$\mathbf{s}_i(t) = UPD(\mathbf{s}'_i(t), \overline{\mathbf{m}}_i(t))$,
where $\mathbf{s}'_i(t)\leftarrow\mathcal{M}_i$. The state update function can also be chosen from different learnable neural networks, e.g., GRU, RNN. Upon computing the new state, the memory of node $i$ is updated by overwriting the corresponding value $\mathcal{M}\leftarrow \mathbf{s}_i(t)$.
Finally, the \textit{Embedding Module} is used to compute the node embedding $\mathbf{emb}_i(t)$ at a specific time $t$:
$\mathbf{emb}_i(t)=\sum_{j\in neighbor_i^k([0,t])} f(\mathbf{s}_i(t), \mathbf{s}_j(t), \mathbf{e}_{ij}(t))$,
where $f$ is a learnable function, e.g., identity, time projection or attention. 
The Decoder $g$ can calculate the probability of the edge existence between two nodes:
$p_{ij}(t) = g(\mathbf{emb}_i(t), \mathbf{emb}_j(t))$, which then provides the self-supervised signals for training.


In our approach to training various TIG models, we first establish a general architecture for most TIG models. This is done by integrating different modules into a single architecture. 
This means all implemented models are specific instances of our approach. 
Moreover, our approach allows these models to be easily extended to accommodate other new models.

\textbf{Distributed Parallel Training.}
Our approach primarily employs multi-GPU computation acceleration to facilitate parallel training of TIG models. To make this possible, the original large graph is divided into several partitions using our \oursp~component. An illustration is in Fig.\ref{fig:flow}(b).

The outputs of \oursp~component are nodes lists $\{\mathcal{V}_1, \dots, \mathcal{V}_p \}$ from which we construct new sub-graphs $\{\mathcal{G}_1, \dots, \mathcal{G}_p \}$ by identifying edges $\mathcal{E}_k = \{(i,j,t) \in \mathcal{E} | i,j \in \mathcal{V}_k \}$ in the original dataset, and we need $\mathcal{N}$ partitions for training. We can choose to divide the original graph to $\mathcal{N}$ partitions directly ($|\mathcal{P}| = \mathcal{N}$). However, after the graph is partitioned, some edges are inevitably deleted. 
Note that we have $\mathcal{V}_a\cup\mathcal{V}_b$ with edge sets being $\mathcal{E}_a\cup\mathcal{E}_b\cup\mathcal{DE}_{ab}$, where $\mathcal{DE}_{ab} = \{e_{ij}(t)\mid i\in\mathcal{V}_a, j\in\mathcal{V}_b\}$ refers to deleted edges between sub-graph $\mathcal{G}_a$ and $\mathcal{G}_b$. We proposed two strategies to relieve the information loss caused by the edge deletion. As outlined in Sec.\ref{sec:TIG part}, shared nodes are added to reduce information loss. 
We can also initially divide the graph into more parts $\{\mathcal{V}_1, \dots, \mathcal{V}_{|\mathcal{P}|} \}, |\mathcal{P}| > \mathcal{N}$. 
Prior to each training epoch, we randomly shuffle all parts and combine them to form $\mathcal{N}$ partitions. When two small partitions are combined, the combined partition will contain the ``deleted'' edges between the two small partitions $combined(\mathcal{V}_a, \mathcal{V}_b)$ with edge sets being $\mathcal{E}_a\cup\mathcal{E}_b\cup\mathcal{DE}_{ab}$. 
Through random shuffling, the ``deleted'' edges between different small partitions can be restored when they are combined, allowing them to be trained across different epochs.

For distributed parallel training based on graph partitioning, we have $\mathcal{N}$ GPUs, and the model will be duplicated into $\mathcal{N}$ copies and deployed on each GPU. The graph data used for training on different GPUs are different sub-graphs which is one of the partitions of the original training graph, i.e., the original graph is been partitioned into $\mathcal{N}$ parts. Assuming we have a partitioning of nodes represented as $\{\mathcal{V}_1, \dots, \mathcal{V}_{\mathcal{N}}\}$, the corresponding interactions, i.e., sub-graphs can be written as $\mathcal{E}_k = \{(i,j,t) \in \mathcal{E} | i,j \in \mathcal{V}_k \}$. Thus we can train sub-graphs parallel at the same time on different GPUs. While only $\mathcal{M}^{(k)} \in \Re^{N^{(k)} \times d}$ memory module is needed for a single-GPU.

In order to balance GPU computational resource utilization and allow for training on larger graphs, our TIG partitioning algorithms ensure that the node counts in each partition are balanced. Based on this setting, we can initialize a memory store module for each GPU with only maximization of all GPUs nodes count. This help us to put graph with very large number of nodes on GPUs.

However, the interaction events, i.e. edges, assigned to different sub-graphs are not exactly the same. Therefore, in order to synchronize the training and the backward of gradients between different GPUs, we design a new training approach. In each epoch, all edges on each GPU are traversed at least once. On GPUs with fewer edges, a loop is made within the epoch. Since the end states of different GPUs may not all be at the end state of a complete data cycle when the entire epoch ends training, this will result in incomplete memory updates of parts of nodes. Thus, we create a backup of the node memory lists at the end of each GPU's data cycle. At the end of the entire epoch, we then restore the node memory across all GPUs to match these latest memory backups. The pseudocode of the training phase is shown in Alg.\ref{alg:training}.


\input{alg_training}

If there are shared nodes, for them, we ensure node memory synchronization across all GPUs. There are two ways of node memory synchronization. The first approach sets the memory value of all shared nodes on every GPU to match memory with the largest timestamp recorded across all GPUs. The second approach resets the memory of all shared nodes by taking an average across all GPUs. After experimental testing, we found that the two synchronization methods have little impact on the performance of downstream tasks, and we adopted the former in our experiments. 


Our distributed parallel acceleration approach allows existing TIG models to be adapted and accelerated. This not only reduces the consumption of computational resources but also significantly decreases computation time. Models which applied our approach also exhibits competitive performance in downstream tasks. 


\section{Experiments}
\label{sec:exps}
\subsection{Datasets, Models, Basic Settings}

We applied \ours~on 7 temporal interaction graph datasets - ML25m \cite{ml25m}, DGraphFin \cite{dgraph}, Taobao \cite{taobao}, Wikipedia, Reddit, MOOC and LastFM \cite{jodie}. Among them, ML25m, DGraphFin and Taobao are three large datasets. 
Some datasets are lack of node features or edge attributes, for which we follow the processing method in previous works \cite{jodie, tgn} by using zero vectors to represent them.
State changes indicators of nodes, i.e., dynamic labels, are included in Wikipedia, Reddit and MOOC, which makes them support dynamic node classification task.
The statistics of all datasets can be found in Tab.\ref{tab:ds}. To avoid information leakage, we chronologically divided the edges into 70\% for training, 15\% for validation, and 15\% for testing before implementing our \oursp.
To prove the efficiency of our methods, we conduct a series of experiments using on 4 different models, i.e., Jodie \cite{jodie}, DyRep \cite{dyrep}, TGN \cite{tgn}, TIGE \cite{tiger}. 

Given the variations between datasets, we apply distinct experimental settings to the large and the small datasets. To optimize training time, we use larger batch sizes for the larger datasets. Specifically, we apply a batch size of 200 to the 4 small datasets, and batch sizes of 2,000, 2,000, and 1,000 are employed when training on ML25m, DGraphFin, and Taobao, respectively. Given that the number of edges in the large datasets exceeds that of the small datasets by over 10 times, we apply smaller maximum training epochs and patience values for the 3 large datasets.

All experiments are conducted on a single server with 72-core CPU, 128GB of memory, and 4 Nvidia Tesla V100 GPUs.


\subsection{Main Experiments}
\input{tab_ds}
\input{tab_time}

\input{tab_main_add_original}

In our experiments, we manipulate the number of shared nodes by adjusting the parameter $top_k$. The choice of $top_k$ is crucial since the proportion of ``important nodes'' varies across real-world datasets. A suitable value of $top_k$ is necessary to strike an optimal balance between edge-cuts across partitions and the workload of the machines in the cluster. 

\subsubsection{Efficiency and computing resources}
To illustrate our advantages in training efficiency and computing resource utilization, we conduct a group of experiments on the 3 big datasets, wherein both training times and the GPU memory allocated by PyTorch modules are recorded. The results are presented in Tab.\ref{tab:time_big}. 

Across all experiments, our methods proved to be the fastest and consumed the fewest GPU resources (on each single GPU). Additionally, the performance of the downstream tasks is highly competitive. By employing our method, training times can be accelerated by up to 8.56x on ML25m, 11.20x on DGraphFin, and 19.27x on Taobao.
In terms of computing resources, our approach is essential for managing large datasets such as DGraphFin and Taobao. Their extensive number of nodes would otherwise lead to out-of-memory (OOM) issues during direct training, particularly due to the storage requirements for extensive node memory (detailed analysis is in \ref{info_and_balancing}).
With the assistance of our method, we distribute the large graph across different GPUs for training to address the OOM problem. Simultaneously, our method alleviates the computational burden on a single GPU and accelerates training.
The efficiency advantages of our partitioning method, compared to the static graph partitioning method, are discussed in Sec.\ref{sec:prat_eff}.

\subsubsection{Performance on down-stream tasks}
\label{sec:pdt}

\textbf{Temporal link prediction.} Following the experimental settings previously outlined in \cite{tiger}, we observe performance on temporal link prediction tasks executed in both transductive and inductive manners. Average precision (AP) score is being used as the evaluation metric. In the case of transductive tasks, the nodes of the predicted edges have appeared during training. Conversely, inductive tasks focus on predicting temporal links between nodes that have not previously been encountered.
The results for different tasks are shown in Tab.\ref{tab:performance}. It is evident that our approach manages to deliver competitive results in downstream tasks with a faster training speed. However, due to OOM issues, some results for larger datasets, that do not applied our method, are unavailable. The best performance on link prediction for both transductive and inductive settings tends to be more focused when $top_k$ is higher. The algorithm HDRF \cite{hdrf} cannot control the number of shared nodes. This may also lead to OOM issues, as excessive node replication and distribution across GPUs occur.
With smaller $top_k$, there are also part of best performance. 
Our method achieves equilibrium by regulating the value of $top_k$. A larger $top_k$ permits more shared nodes, thereby collating additional information beneficial to downstream tasks. Conversely, a smaller $top_k$ helps filter out potential noise or irrelevant edges, further enhancing downstream task performance. Overall, managing $top_k$ guarantees both acceleration on large datasets and advantages for downstream tasks.

\input{tab_nodes}

\input{tab_cut}

\textbf{Node classification.} Considering node classification tasks require dynamic labels, we conduct our experiments on datasets with available labels, i.e., Wikipedia, Reddit and MOOC. The performances evaluated by AUROC (area under the ROC curve) are shown in Tab.\ref{tab:node_class}, as the same as the previous works used \cite{jodie, dyrep, tgat, tgn, tiger}. 
As the results show, the application of our methods can yield results that surpass those achieved using the original models without partitioning, across various models and datasets. This further underscores the effectiveness of \ours~in enhancing the performance of downstream tasks.

\begin{figure}[!tbp]
	\centering
	\subfigbottomskip=2pt
	\subfigcapskip=-5pt
	\subfigure{\includegraphics[width=.45\columnwidth]{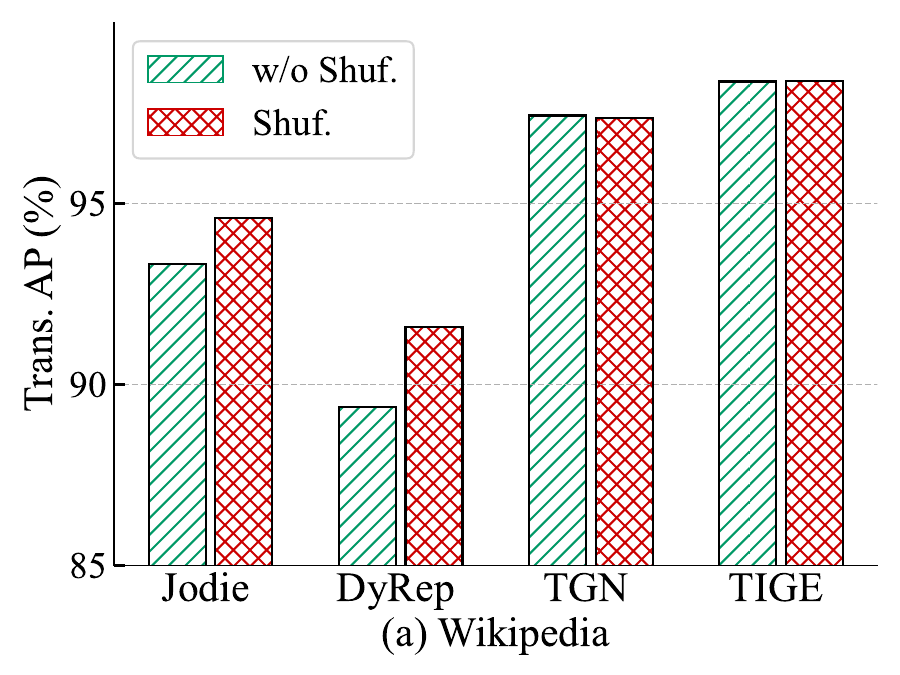}}\hspace{5pt}
	\subfigure{\includegraphics[width=.45\columnwidth]{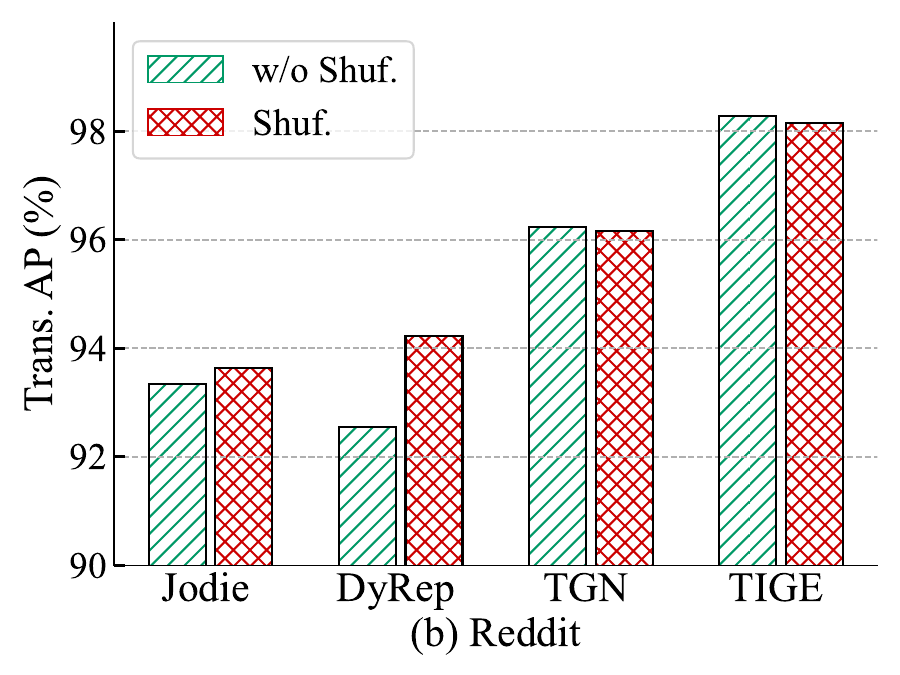}}\\
	\subfigure{\includegraphics[width=.45\columnwidth]{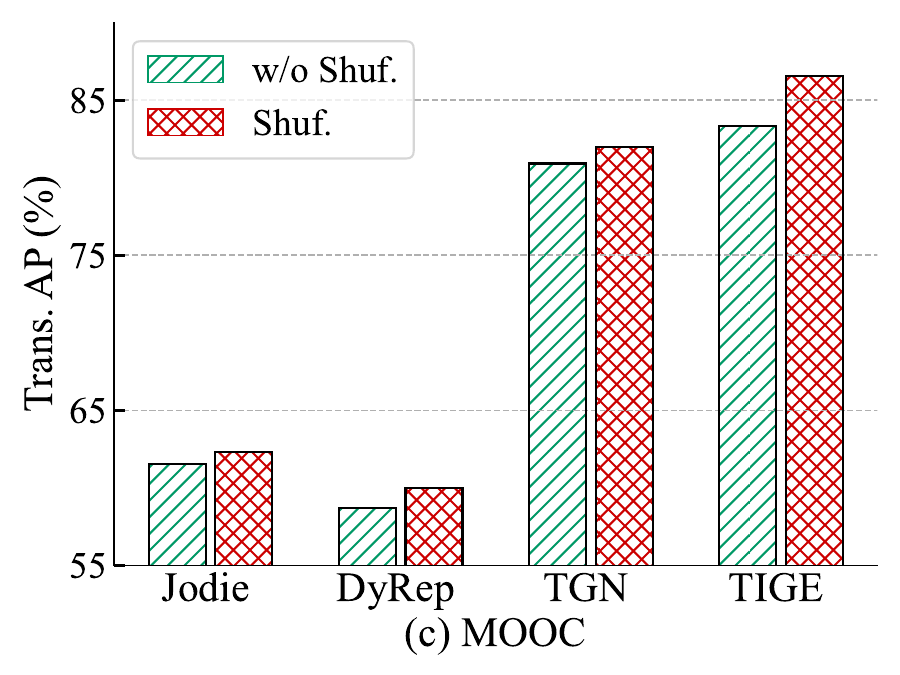}}\hspace{5pt}
	\subfigure{\includegraphics[width=.45\columnwidth]{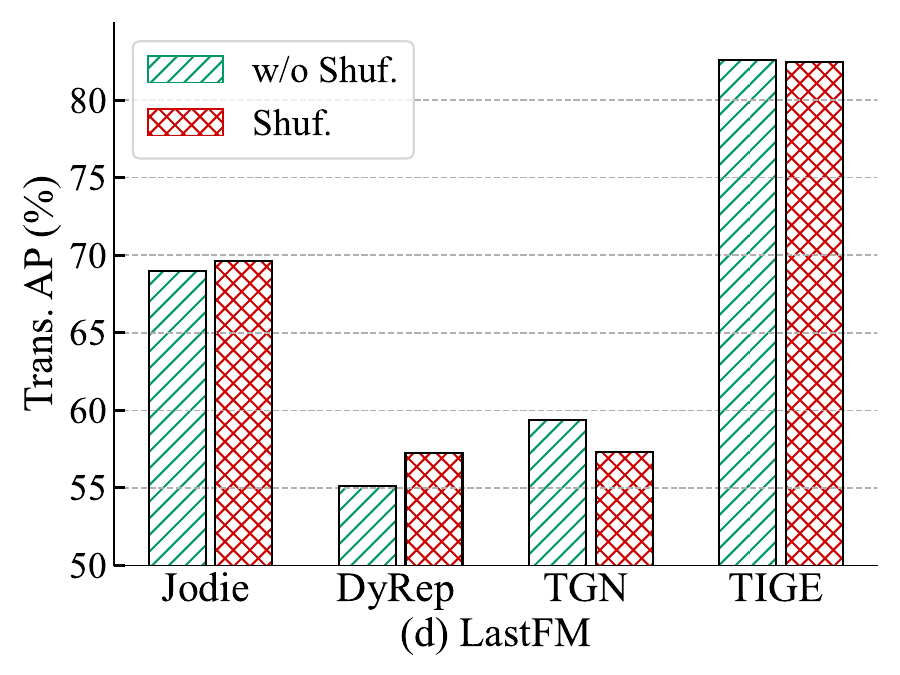}}
\caption{Impacts of partition shuffling.}
\label{fig:shuffle}
\end{figure}

\begin{figure}[!tbp]
	\centering
	\subfigbottomskip=2pt
	\subfigcapskip=-5pt
	\subfigure{\includegraphics[width=.45\columnwidth]{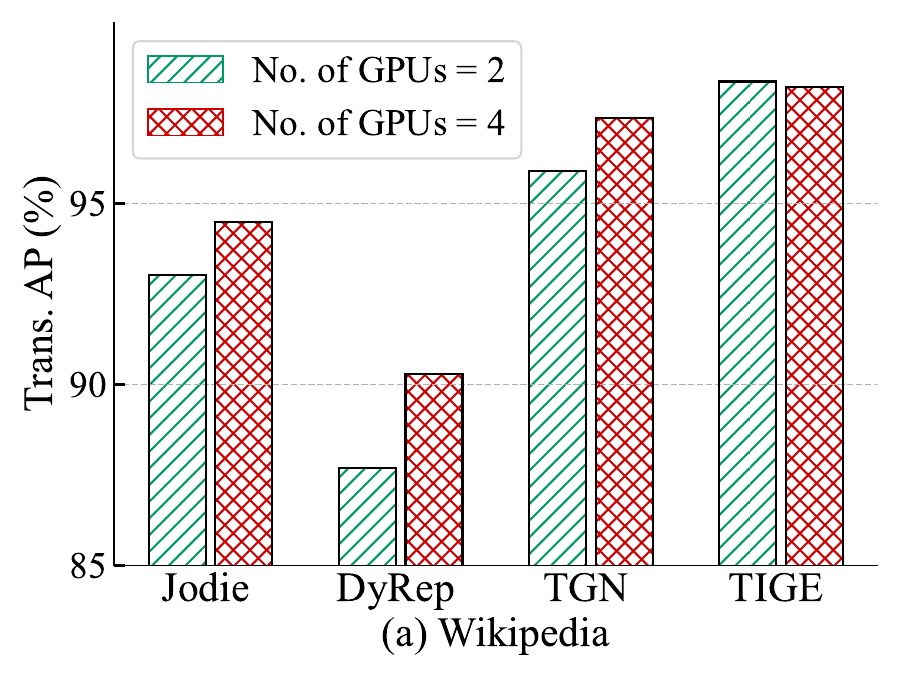}}\hspace{5pt}
	\subfigure{\includegraphics[width=.45\columnwidth]{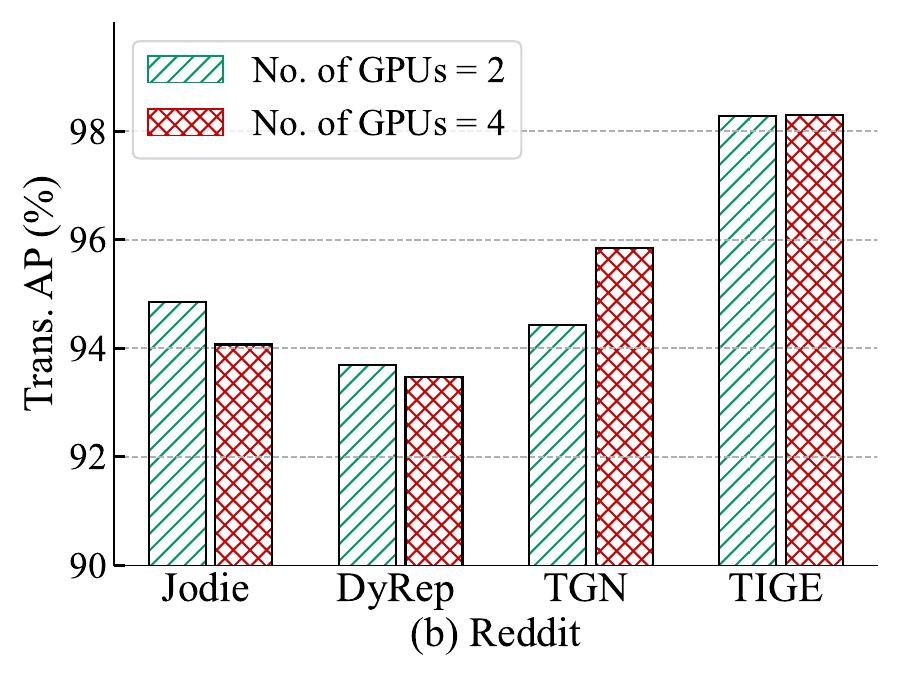}}\\
	\subfigure{\includegraphics[width=.45\columnwidth]{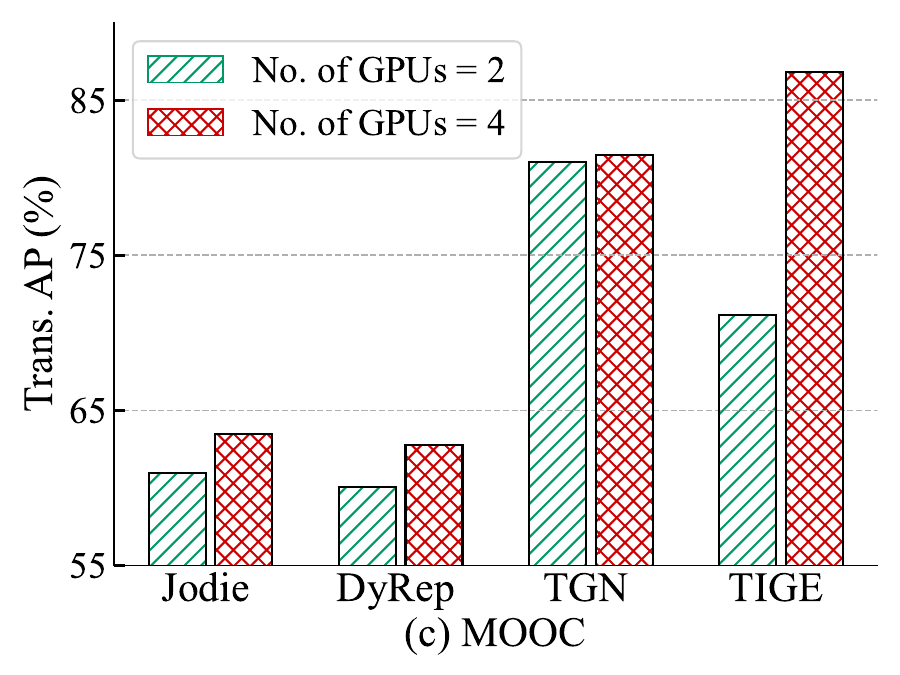}}\hspace{5pt}
	\subfigure{\includegraphics[width=.45\columnwidth]{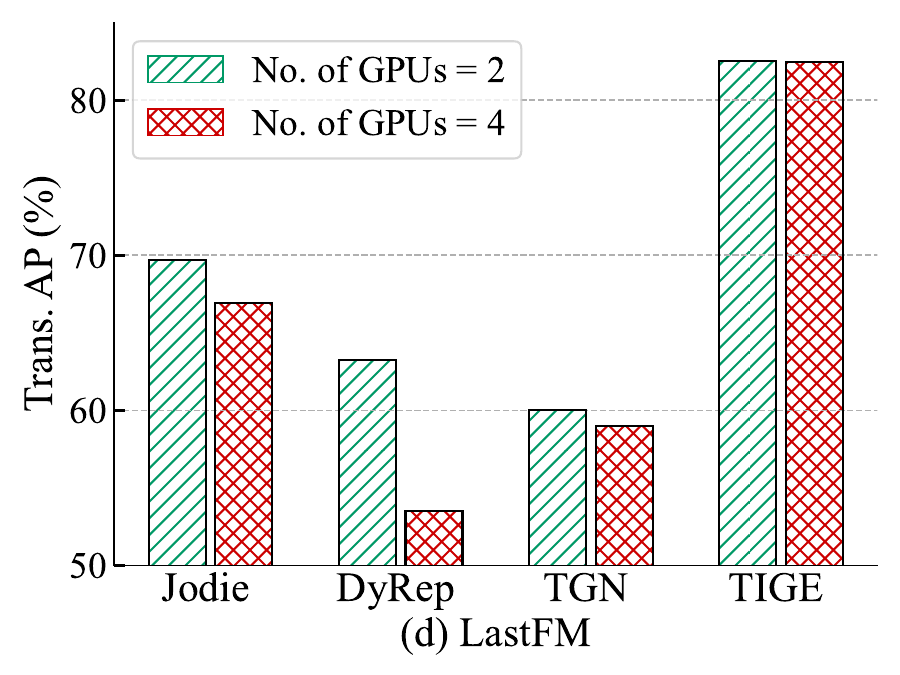}}
\caption{Impacts of Changing the Number of GPUs ($\mathcal{N}$).}
\label{fig:ws}
\end{figure}

\subsubsection{Information loss and load balancing analysis}
\label{info_and_balancing}
As the TIG partitioning deletes edges, and information loss occurs, the performance of downstream tasks will inevitably be affected. While adding shared nodes to different sub-graphs can alleviate some of the information loss caused by edge deletion, synchronizing node memory will also result in information loss from the shared nodes. Simultaneously, the imbalance in the distribution of edges and nodes can slow down the entire training process and impact GPU resource allocation, respectively.

\input{tab_kl}
\input{tab_par_speedup}

Thus, we choose the largest dataset Taobao as an example for further investigation (statistics are shown in Tab.\ref{tab:cut}). we compute the total number of deleted edges (shown as ``total cut'' in the table), the edges on different sub-graphs, i.e, different GPUs (shown as the ``edges std.'' in the table) and the nodes on different sub-graphs (shown as the ``nodes std.'' in the table).
We can find that as the number of shared nodes increases ($top_k$), edge deletion decreases (edge cut). 
Our method is balancing the load in terms of both the number of edges and nodes. The balance of the number of edges enables our method to have a better acceleration effect and speed up the overall training time. At the same time, the balance of the number of nodes enables our method to better balance the use of different GPU resources, so that graphs with a larger number of nodes can be accelerated for training.

Compared to the KL method, due to the imbalanced distribution of edges, training speed is slower than with graphs partitioned by our \oursp. 
Moreover, HDRF does not control the number of shared nodes, which results in a large number of nodes being distributed on GPUs. This results that although the graph with large number of nodes is partitioned by HDRF, as shown in Tab.\ref{tab:time_big} and Tab.\ref{tab:performance}, it is not feasible to train it on a multi-GPU setup.

\subsection{Comparative Experiments}
We also conduct two sets of comparative experiments to demonstrate the effect of our components or different experimental settings on down-stream tasks. 

\subsubsection{Shuffle Partitions}
\label{sec:sp}

This set of experiments is designed to investigate the effects of the shuffle partitions method on various datasets and models. Specifically, all graphs are initially partitioned into eight small partitions. During training, these eight partitions are shuffled and combined into 4 partitions. Alternatively, without shuffling, they are directly combined into 4 partitions. These combined 4 partitions are then trained on 4 GPUs in parallel.
We present the results on the 4 datasets by setting $top_k = 5$, as shown in Fig.\ref{fig:shuffle}. similar trend holds for the other datasets which cannot be shown due to space limitation.
The results of the experiments illustrate that the shuffle partitions is effective in the majority of cases.

\subsubsection{Change of Number of GPUs ($\mathcal{N}$)}
To delve deeper into the impact of changing the number of partitions or GPUs (i.e., $\mathcal{N}$) on downstream tasks, we conduct another set of experiments. (The results are shown in Fig.\ref{fig:ws}.) The original graph is partitioned into either 2 or 4 parts, with the corresponding number of GPUs (2 or 4) being used directly for training. The results illustrate the impact of changing number of partitions and number of GPU cards. Meanwhile, comparing the experimental results in shuffle partitions, the impact of partitioning graph into more parts and then randomly splicing them or directly partition the graph into number of parts equals to $\mathcal{N}$.

An increase in the $\mathcal{N}$ leads to an increase in edge deletions, as the $\mathcal{N}$ should correspond to the number of sub-graphs. Thus, an increase in the number of sub-graphs results in an increased number of deleted edges, implying more information loss. As previously mentioned, the effectiveness of model with deleted edges might be impacted. This is because the deleted edges could potentially contain noise which the deletion of them can contribute to the performance.

\subsection{Compare with Static Graph Partitioning Algorithm}
\label{sec:kl}
In the realm of static graph partitioning algorithms, both KL\cite{kl} and HDRF\cite{hdrf} are considered representative approaches. However, as HDRF is a special case within our approach, we mainly use KL as the representative method for comparison.
Static graph partitioning algorithms generally achieve low edge cuts because they can access global graph information. However, a competent partitioning algorithm should not only deliver quality results but also be time-efficient and achieve load balancing. KL\cite{kl} is a representative algorithm for static graph partitioning. Tab.\ref{tab:cut} demonstrates that the KL performs well on edge cuts without shared nodes, but performs worst on load balancing, i.e., standard deviation of edges. Furthermore, we train models using our proposed \oursd~to compare the speed-up of training time and performance in downstream tasks. We also evaluate partitioning time across different datasets. The detailed analysis is as follows.

\subsubsection{Performance and Training Time Speed-up on Downstream Tasks}
As presented in Tab.\ref{tab:kl}, training times using the KL algorithm  are comparatively longer than those of our method (for which we present the results for $top_k=0$, since KL also does not use shared nodes). This discrepancy is more pronounced in datasets with a larger number of edges. The reason for this is that KL ignores edge balancing, resulting in an uneven distribution of edges across different GPUs. Our training approach will loop over epochs for GPUs with fewer edge data, resulting in these data portions being trained more times compared to other GPUs with more edge data. While, GPUs with more edges may trained only one cycle and took longer time. This leads to two problems, the first one is the training time increases, and the second is the unbalanced data traversing. The second problem will also case the uncertainty of downstream task performances. Our methods accelerate training up to 7.2x on ML25m, 1.1x on DGraphFin, and 10.7x on Taobao compared to KL. Our methods with $top_k=0$ outperform other approaches on most datasets and models. Note that increasing $top_k$ from 0 to a higher value might enhance performance due to information loss reduced.

\subsubsection{Efficiency on Graph partitioning}
\label{sec:prat_eff}
As is evident from the Tab.\ref{tab:par_speedup}, the efficiency advantage of our \oursp~over the KL becomes more pronounced as the size of the dataset increases. \oursp~can enhance the graph partitioning speed by up to a factor of 94.57x compared to the KL algorithm. In scenarios where real-time performance is required, especially when the graph is dynamically changing, the additional overhead associated with the re-partitioning of the KL algorithm is not feasible.

\balance

\section{Related Work}
\textbf{TIG Embedding.} TIG models capture the dynamic nature of graphs, thereby enabling superior modelling of TIGs. Jodie \cite{jodie} employs two Recurrent Neural Networks (RNNs) to dynamically update node representations. DyRep \cite{dyrep} proposes a deep temporal point process model that utilizes a two-time scale approach to capture both association and communication information. TGN \cite{tgn} introduces a memory-based approach for TIG embedding. TIGE \cite{tiger} puts forward a model that incorporates a dual-memory module for more effective aggregation of neighbour information. Given that most existing models are constrained to single-GPU training, there exists a compelling motivation for the proposal of a distributed training approach for TIG models.

\textbf{GNN Training Acceleration.} 
For static Graph Neural Networks (GNNs), numerous studies \cite{sage, vrgcn, fastgcn, clustergcn} have attempted to implement large graph sampling for training. However, as the graph size and the number of model layers expand, they invariably encounter the ``neighborhood explosion'' problem.

Efforts have been made to achieve distributed full batch training \cite{neugraph,roc,dgcl,distgnn}, but these often compromise model convergence and accuracy. Distributed GNN mini-batch training represents an alternative platforms like AliGraph \cite{aligraph} and AGL \cite{agl}, though industrial-scale, do not utilize GPU acceleration.
DistDGL \cite{distdgl} employs synchronized stochastic gradient descent for distributed training and maintains a Key-Value store to efficiently acquire graph information. BGL \cite{bgl} introduces a dynamic caching mechanism to minimize CPU-GPU communication overhead. ByteGNN \cite{bytegnn} enables the mini-batch sampling phase to be parallelizable by viewing it as a series of Directed Acyclic Graphs with small tasks. 

A handful of studies have concentrated on accelerating temporal GNNs training. EDGE \cite{edge} improves computational parallelism by selecting and duplicating specific $d$-nodes, thereby eliminating certain computational dependencies. However, its applicability is confined to Jodie \cite{jodie}, limiting generalizability. TGL \cite{tgl} introduces a Temporal-CSR data structure, coupled with a parallel sampler, to sample neighboring nodes efficiently for mini-batch training. However, it is not tailored for distributed training and thus orthogonal to our work.

\textbf{Graph Partitioning in GNNs.} METIS \cite{metis} is a multi-stage static partitioning method designed to minimize edge cuts. It is used by \cite{clustergcn} to construct a batch during training and by \cite{dgcl, aligraph, distdgl} to partition large graphs for distributed training. NeuGraph \cite{neugraph} utilizes KL \cite{kl} to maximize the assignment of edges connected to the same node into the same partition. However, such static graph partitioning methods have high time complexity and require re-partitioning when the graph changes. Euler \cite{euler} and Roc \cite{roc} apply methods such as random partitioning and linear regression-based techniques. They ignore the graph structural information, resulting in lower quality of partitioning as well as unbalanced computational load.

Streaming graph partitioning methods aim to perceive the graph as an edge-stream or node-stream input. AliGraph \cite{aligraph} incorporates Linear Deterministic Greedy (LGD) \cite{lgd}, an edge-cut based method suited for partitioning dynamically evolving graphs. DistGNN \cite{distgnn} uses a node-cut based method Libra \cite{libra}. However, it relies on a hash function to randomly assign edges, thereby ignoring the structural information of the graph and resulting in a high edge-cut ratio. Greedy \cite{greedy} and HDRF \cite{hdrf} have been shown to have better partitioning quality \cite{sgpstudy}. However, they either only suitable for static graphs or regard edges at different timestamps equivalently, failing to utilize the characteristics of temporal interaction graphs. Also, they face an excessive number of replica nodes when partitioning real-world graph data. This insight drives us to propose a novel partitioning method tailored for TIGs.

\section{Conclusion}
In this paper, we propose a novel Temporal Interaction Graph embedding approach consisting of a streaming edge partitioning method, accompanied by a corresponding distributed parallel training component. By applying our approach, we can efficiently train very large-scale temporal interaction graphs on GPUs. Moreover, our approach can be accelerated using distributed parallel training with multiple GPUs.
Our experiments demonstrate that our methods can handle TIGs with millions of nodes and billions of edges. In contrast, previous methods are unable to directly train such large graphs due to computing resource limitations. 
In future work, we intend to further investigate the impact of edge deletion and strive to provide more interpretability to the information loss issue, concentrating on eliminating noisy or unimportant edges while retaining valid ones. 

\newpage
\bibliographystyle{unsrt}  
\bibliography{references}

\end{document}

%% file: tab_1.tex
\begin{table}[t]
\centering
\caption{Comparison of different graph partition
algorithms}
\setlength{\tabcolsep}{1.5mm}{
\begin{tabular}{c|c|c|c|c}
\toprule
\begin{tabular}[c]{@{}c@{}}\textbf{Patition} \\ \textbf{Algorithms}\end{tabular} & \begin{tabular}[c]{@{}c@{}}\textbf{Support Time} \\ \textbf{Information}\end{tabular}  & \begin{tabular}[c]{@{}c@{}}\textbf{Low Repli-} \\ \textbf{cation Factor}\end{tabular}  & \begin{tabular}[c]{@{}c@{}}\textbf{Load} \\ \textbf{Balance}\end{tabular}  & \textbf{Scalable} \\ \midrule
\begin{tabular}[c]{@{}c@{}} METIS\cite{metis} \& \\ KL\cite{kl} \end{tabular}  & \XSolidBrush  &\CheckmarkBold  &\XSolidBrush  &\XSolidBrush \\
\midrule
\begin{tabular}[c]{@{}c@{}} Random\cite{euler} \& \\ LGD\cite{lgd} \end{tabular}  & \XSolidBrush  &\CheckmarkBold  &\XSolidBrush  &\CheckmarkBold \\
\midrule
ROC\cite{roc}  & \XSolidBrush  &\XSolidBrush  &\XSolidBrush  &\CheckmarkBold \\
\midrule
\begin{tabular}[c]{@{}c@{}} Libra\cite{libra} \& \\ Greedy\cite{greedy} \& \\ HDRF\cite{hdrf} \end{tabular}  &\XSolidBrush  &\XSolidBrush  &\CheckmarkBold  &\CheckmarkBold \\
\midrule
Ours  &\CheckmarkBold  &\CheckmarkBold  &\CheckmarkBold  &\CheckmarkBold \\
\bottomrule
\end{tabular}
}
\label{tab:1}
\end{table}

%% file: alg_partitioning.tex
\begin{algorithm}[t]
\label{alg:partitioning}
\caption{Streaming Edge Partitioning}
    \KwIn{Edge set $\mathcal{E}$, Node set $\mathcal{V}$, Proportion of replicable nodes $k$}
    \KwOut{Partition of nodes $\mathcal{P} = (p_1,...,p_n)$ and a shared nodes list $\mathcal{S}$}
    Scan the input edge set $\mathcal{E}$ to calculate the centrality value $Cent(i)$ of each node $i$ in $\mathcal{V}$. Return the set of nodes $\mathcal{T}$ with the first $k*|\mathcal{V}|$ largest centrality value\;
    \For{each edge e = (i,j,t) in $\mathcal{E}$}{
        $A(i)$ is the set of partitions to which node $i$ has been assigned\;
        \If{$A(i) \neq \varnothing$ and $A(j) \neq \varnothing$}{
            \textbf{Case 1: }$i(j) \in \mathcal{T}$ and $j(i) \notin \mathcal{T}$\;
                Assign $e$ to the partition $\hat{p}$ where $j(i)$ resides\;
            \textbf{Case 2: }$i \in \mathcal{T}$ and $j \in \mathcal{T}$\;
                Assign $e$ to the partition $\hat{p}$ with maximum score $C(i,j,p)$\;
            \textbf{Case 3: }$i \notin \mathcal{T}$ and $j \notin \mathcal{T}$\;
                \If{$A(i)==A(j)=={\hat{p}}$}{
                Assign $e$ to the partition $\hat{p}$\;
                }
                \Else{Discard the edge $e$\;}
        }
        \Else{
            \textbf{Case 4 \& 5: }either both nodes are unassigned or one of the nodes is assigned\;
                Assign $e$ to the partition $\hat{p}$ with maximum score $C(i,j,p)$\;
        }
    }
    \For{each node $i \in \mathcal{V}$}
    {
    \If{$|A(i)|$ $>$ 1}{
        Add $i$ to the shared nodes list $\mathcal{S}$\;
        Add $i$ to all the partitions in $\mathcal{P}$\;
    }
    \Else{
        Add $i$ to the $p \in A(i)$
    }
    }
\end{algorithm}

%% file: alg_training.tex
\begin{algorithm}[!t]
\label{alg:training}
\caption{Training Loop}
    \KwData{Test Dataloader}
    $i\_batch = 0$\;
    \While{True}{
        \For{i, data in enumerate(Test Dataloader)}{
        $loop\_start = i == 0$\;
        $loop\_end = i == len(dataloader) - 1$\;
        \If{$loop\_start$}{reset nodes memory\;}
        \textbf{Training and compute loss}\;
        Backward and Optimize\;
        \If{$loop\_end$}{backup current nodes memory state\;}
        $i\_batch += 1$\;
        }
    \If{All GPUs have finished at least one complete traversal for all mini batches}{break}
    }
\end{algorithm}

%% file: tab_ds.tex
\begin{table}[t]
\centering
\caption{Dataset Statistic. $d_{n}$ and $d_{e}$ indicate the dim of nodes and edges, respectively. Classes means the numbers of labels.}
\begin{tabular}{c|ccccc}
\toprule
          & \# Nodes  & \# Edges    & $d_{n}$ & $d_{e}$ & Classes \\ \midrule
Wikipedia & 9,227     & 157,474     & 172           & 172           &    2    \\
Reddit    & 10,984    & 672,447     & 172           & 172           &    2    \\
MOOC      & 7,144     & 411,749     & 172           & 172           &    2    \\
LastFM    & 1,980     & 1,293,103   & 172           & 172           &    -    \\ \midrule
ML25m     & 221,588   & 25,000,095  & 100           & 1             &    -    \\
DGraphFin & 4,889,537 & 4,300,999   & 100           & 11            &    4    \\
Taobao    & 5,149,747 & 100,135,088 & 100           & 4             &    9,439  \\ \bottomrule
\end{tabular}
\label{tab:ds}
\end{table}


%% file: tab_time.tex
\begin{table*}[!htb]
\centering
\caption{Running (training) times (per epoch in seconds) and GPU memory resources reserved (per GPU in GB) for 3 big datasets. Uses training on CPU as speed comparison baseline. (Results are based on self-supervised training of link prediction. All our methods are trained on a 4x Nvidia Tesla V100 machine and graphs are divided into 4 partitions.)}
\begin{tabular}{ccccccccccc}
\toprule
                                                                      &                                 & \multicolumn{3}{c}{Ml25m}                                                                                                                                                                         & \multicolumn{3}{c}{DGraphFin}                                                                                                                                                                     & \multicolumn{3}{c}{Taobao}                                                                                                                                                   \\ \midrule
                                                                      &                                 & \begin{tabular}[c]{@{}c@{}}Training\\  Time\end{tabular} & \begin{tabular}[c]{@{}c@{}}Speed\\ -up\end{tabular} & \multicolumn{1}{c|}{\begin{tabular}[c]{@{}c@{}}GPU Mem.\\ Reserved\end{tabular}} & \begin{tabular}[c]{@{}c@{}}Training\\  Time\end{tabular} & \begin{tabular}[c]{@{}c@{}}Speed\\ -up\end{tabular} & \multicolumn{1}{c|}{\begin{tabular}[c]{@{}c@{}}GPU Mem.\\ Reserved\end{tabular}} & \begin{tabular}[c]{@{}c@{}}Training\\  Time\end{tabular} & \begin{tabular}[c]{@{}c@{}}Speed\\ -up\end{tabular} & \begin{tabular}[c]{@{}c@{}}GPU Mem.\\ Reserved\end{tabular} \\ \midrule
\multicolumn{1}{c|}{\multirow{7}{*}{\rotatebox[origin=c]{90}{Jodie}}} & \multicolumn{1}{c|}{$top_k=0$}  & 226.50                                                   & \textbf{5.90x}                                      & \multicolumn{1}{c|}{\textbf{0.88}}                                               & 102.70                                                   & \textbf{5.00x}                                      & \multicolumn{1}{c|}{\textbf{10.19}}                                              & 917.29                                                   & \textbf{8.88x}                                      & \textbf{12.25}                                              \\
\multicolumn{1}{c|}{}                                                 & \multicolumn{1}{c|}{$top_k=1$}  & 414.73                                                   & 3.22x                                               & \multicolumn{1}{c|}{0.91}                                                        & 105.02                                                   & 4.89x                                               & \multicolumn{1}{c|}{13.44}                                                       & 2096.44                                                  & 3.89x                                               & 16.59                                                       \\
\multicolumn{1}{c|}{}                                                 & \multicolumn{1}{c|}{$top_k=5$}  & 773.50                                                   & 1.73x                                               & \multicolumn{1}{c|}{1.02}                                                        & 117.47                                                   & 4.37x                                               & \multicolumn{1}{c|}{14.72}                                                       & 4866.02                                                  & 1.67x                                               & 19.05                                                       \\
\multicolumn{1}{c|}{}                                                 & \multicolumn{1}{c|}{$top_k=10$} & 1045.46                                                  & 1.28x                                               & \multicolumn{1}{c|}{1.17}                                                        & 134.26                                                   & 3.82x                                               & \multicolumn{1}{c|}{15.98}                                                       & 7326.25                                                  & 1.11x                                               & 24.62                                                       \\
\multicolumn{1}{c|}{}                                                 & \multicolumn{1}{c|}{HDRF}       & 1177.96                                                  & 1.13x                                               & \multicolumn{1}{c|}{1.90}                                                        & OOM                                                      & OOM                                                 & \multicolumn{1}{c|}{OOM}                                                         & OOM                                                      & OOM                                                 & OOM                                                         \\
\multicolumn{1}{c|}{}                                                 & \multicolumn{1}{c|}{Single-GPU} & 1313.66                                                  & 1.02x                                               & \multicolumn{1}{c|}{2.82}                                                        & OOM                                                      & OOM                                                 & \multicolumn{1}{c|}{OOM}                                                         & OOM                                                      & OOM                                                 & OOM                                                         \\
\multicolumn{1}{c|}{}                                                 & \multicolumn{1}{c|}{CPU}        & 1335.61                                                  & 1x                                                  & \multicolumn{1}{c|}{-}                                                           & 513.30                                                   & 1x                                                  & \multicolumn{1}{c|}{-}                                                           & 8147.59                                                  & 1x                                                  & -                                                           \\ \midrule
\multicolumn{1}{c|}{\multirow{7}{*}{\rotatebox[origin=c]{90}{Dyrep}}} & \multicolumn{1}{c|}{$top_k=0$}  & 236.99                                                   & \textbf{7.12x}                                      & \multicolumn{1}{c|}{\textbf{1.25}}                                               & 105.18                                                   & \textbf{7.69x}                                      & \multicolumn{1}{c|}{\textbf{10.20}}                                              & 961.75                                                   & \textbf{15.64x}                                     & \textbf{12.26}                                              \\
\multicolumn{1}{c|}{}                                                 & \multicolumn{1}{c|}{$top_k=1$}  & 431.77                                                   & 3.91x                                               & \multicolumn{1}{c|}{1.28}                                                        & 107.37                                                   & 7.54x                                               & \multicolumn{1}{c|}{12.92}                                                       & 2219.06                                                  & 6.78x                                               & 16.13                                                       \\
\multicolumn{1}{c|}{}                                                 & \multicolumn{1}{c|}{$top_k=5$}  & 812.03                                                   & 2.08x                                               & \multicolumn{1}{c|}{1.45}                                                        & 121.85                                                   & 6.64x                                               & \multicolumn{1}{c|}{14.57}                                                       & 5186.08                                                  & 2.90x                                               & 20.79                                                       \\
\multicolumn{1}{c|}{}                                                 & \multicolumn{1}{c|}{$top_k=10$} & 1095.72                                                  & 1.54x                                               & \multicolumn{1}{c|}{1.61}                                                        & 136.58                                                   & 5.92x                                               & \multicolumn{1}{c|}{15.90}                                                       & 7964.41                                                  & 1.89x                                               & 24.62                                                       \\
\multicolumn{1}{c|}{}                                                 & \multicolumn{1}{c|}{HDRF}       & 1238.93                                                  & 1.36x                                               & \multicolumn{1}{c|}{2.30}                                                        & OOM                                                      & OOM                                                 & \multicolumn{1}{c|}{OOM}                                                         & OOM                                                      & OOM                                                 & OOM                                                         \\
\multicolumn{1}{c|}{}                                                 & \multicolumn{1}{c|}{Single-GPU} & 1403.57                                                  & 1.20x                                               & \multicolumn{1}{c|}{3.42}                                                        & OOM                                                      & OOM                                                 & \multicolumn{1}{c|}{OOM}                                                         & OOM                                                      & OOM                                                 & OOM                                                         \\
\multicolumn{1}{c|}{}                                                 & \multicolumn{1}{c|}{CPU}        & 1687.00                                                  & 1x                                                  & \multicolumn{1}{c|}{-}                                                           & 809.19                                                   & 1.00                                                & \multicolumn{1}{c|}{-}                                                           & 15045.97                                                 & 1x                                                  & -                                                           \\ \midrule
\multicolumn{1}{c|}{\multirow{7}{*}{\rotatebox[origin=c]{90}{TGN}}}   & \multicolumn{1}{c|}{$top_k=0$}  & 238.35                                                   & \textbf{11.13x}                                     & \multicolumn{1}{c|}{\textbf{1.26}}                                               & 107.49                                                   & \textbf{8.56x}                                      & \multicolumn{1}{c|}{\textbf{11.89}}                                              & 976.55                                                   & \textbf{18.83x}                                     & \textbf{14.08}                                              \\
\multicolumn{1}{c|}{}                                                 & \multicolumn{1}{c|}{$top_k=1$}  & 441.81                                                   & 6.00x                                               & \multicolumn{1}{c|}{1.27}                                                        & 111.95                                                   & 8.22x                                               & \multicolumn{1}{c|}{14.98}                                                       & 2266.76                                                  & 8.11x                                               & 18.04                                                       \\
\multicolumn{1}{c|}{}                                                 & \multicolumn{1}{c|}{$top_k=5$}  & 821.24                                                   & 3.23x                                               & \multicolumn{1}{c|}{1.46}                                                        & 126.33                                                   & 7.29x                                               & \multicolumn{1}{c|}{16.35}                                                       & 5312.83                                                  & 3.46x                                               & 21.69                                                       \\
\multicolumn{1}{c|}{}                                                 & \multicolumn{1}{c|}{$top_k=10$} & 1109.66                                                  & 2.39x                                               & \multicolumn{1}{c|}{1.64}                                                        & 140.46                                                   & 6.55x                                               & \multicolumn{1}{c|}{17.84}                                                       & 8058.26                                                  & 2.28x                                               & 25.50                                                       \\
\multicolumn{1}{c|}{}                                                 & \multicolumn{1}{c|}{HDRF}       & 1255.80                                                  & 2.11x                                               & \multicolumn{1}{c|}{2.33}                                                        & OOM                                                      & OOM                                                 & \multicolumn{1}{c|}{OOM}                                                         & OOM                                                      & OOM                                                 & OOM                                                         \\
\multicolumn{1}{c|}{}                                                 & \multicolumn{1}{c|}{Single-GPU} & 1358.06                                                  & 1.95x                                               & \multicolumn{1}{c|}{3.92}                                                        & OOM                                                      & OOM                                                 & \multicolumn{1}{c|}{OOM}                                                         & OOM                                                      & OOM                                                 & OOM                                                         \\
\multicolumn{1}{c|}{}                                                 & \multicolumn{1}{c|}{CPU}        & 2652.76                                                  & 1x                                                  & \multicolumn{1}{c|}{-}                                                           & 920.37                                                   & 1x                                                  & \multicolumn{1}{c|}{-}                                                           & 18389.03                                                 & 1x                                                  & -                                                           \\ \midrule
\multicolumn{1}{c|}{\multirow{7}{*}{\rotatebox[origin=c]{90}{TIGE}}}  & \multicolumn{1}{c|}{$top_k=0$}  & 251.35                                                   & \textbf{11.20x}                                     & \multicolumn{1}{c|}{\textbf{1.26}}                                               & 108.10                                                   & \textbf{8.00x}                                      & \multicolumn{1}{c|}{\textbf{11.89}}                                              & 984.37                                                   & \textbf{19.27x}                                     & \textbf{14.08}                                              \\
\multicolumn{1}{c|}{}                                                 & \multicolumn{1}{c|}{$top_k=1$}  & 443.86                                                   & 6.34x                                               & \multicolumn{1}{c|}{1.27}                                                        & 112.56                                                   & 7.69x                                               & \multicolumn{1}{c|}{14.98}                                                       & 2273.37                                                  & 8.34x                                               & 18.04                                                       \\
\multicolumn{1}{c|}{}                                                 & \multicolumn{1}{c|}{$top_k=5$}  & 825.69                                                   & 3.41x                                               & \multicolumn{1}{c|}{1.44}                                                        & 125.76                                                   & 6.88x                                               & \multicolumn{1}{c|}{16.35}                                                       & 5357.19                                                  & 3.54x                                               & 21.69                                                       \\
\multicolumn{1}{c|}{}                                                 & \multicolumn{1}{c|}{$top_k=10$} & 1109.07                                                  & 2.54x                                               & \multicolumn{1}{c|}{1.63}                                                        & 142.37                                                   & 6.08x                                               & \multicolumn{1}{c|}{17.85}                                                       & 8110.54                                                  & 2.34x                                               & 25.50                                                       \\
\multicolumn{1}{c|}{}                                                 & \multicolumn{1}{c|}{HDRF}       & 1231.67                                                  & 2.28x                                               & \multicolumn{1}{c|}{2.31}                                                        & OOM                                                      & OOM                                                 & \multicolumn{1}{c|}{OOM}                                                         & OOM                                                      & OOM                                                 & OOM                                                         \\
\multicolumn{1}{c|}{}                                                 & \multicolumn{1}{c|}{Single-GPU} & 1284.87                                                  & 2.19x                                               & \multicolumn{1}{c|}{3.92}                                                        & OOM                                                      & OOM                                                 & \multicolumn{1}{c|}{OOM}                                                         & OOM                                                      & OOM                                                 & OOM                                                         \\
\multicolumn{1}{c|}{}                                                 & \multicolumn{1}{c|}{CPU}        & 2813.96                                                  & 1x                                                  & \multicolumn{1}{c|}{-}                                                           & 865.23                                                   & 1x                                                  & \multicolumn{1}{c|}{-}                                                           & 18967.77                                                 & 1x                                                  & -                                                           \\ \bottomrule
\end{tabular}

\label{tab:time_big}
\end{table*}

%% file: tab_main_add_original.tex
\useunder{\uline}{\ul}{}
\begin{table*}[!t]
\centering
\caption{Average Precision (\%) for future edge prediction task in transductive and inductive settings. $top_k$ refers to the hyper-parameter that control the percentage of shared nodes. When there is no restriction for $top_k$ the algorithm degenerates to HDRF. The best AP of different $top_k$ on different datasets and backbones are in bold. Parts of the results are unavailable for ML25m, due to the original methods cannot efficiently process the dataset directly.}

\begin{tabular}{ccccccc|ccc}
\toprule
                                                                              &                                                                       &                                       & Wikipeida          & Reddit             & MOOC               & LastFM             & ML25m              & DGraphFin          & Taobao             \\ \midrule
\multicolumn{1}{c|}{\multirow{24}{*}{\rotatebox[origin=c]{90}{Transductive}}} & \multicolumn{1}{c|}{\multirow{6}{*}{\rotatebox[origin=c]{90}{Jodie}}} & \multicolumn{1}{c|}{$top_k=0$}        & 94.87±0.7          & 94.13±0.6          & 61.54±0.8          & 65.12±5.1          & 92.22±2.1          & 74.75±1.7          & 91.43±1.1          \\
\multicolumn{1}{c|}{}                                                         & \multicolumn{1}{c|}{}                                                 & \multicolumn{1}{c|}{$top_k=1$}        & 91.12±0.3          & 90.29±1.7          & 61.87±1.2          & 67.69±0.9          & 93.70±0.5          & \textbf{74.77±0.6} & 91.28±0.6          \\
\multicolumn{1}{c|}{}                                                         & \multicolumn{1}{c|}{}                                                 & \multicolumn{1}{c|}{$top_k=5$}        & 94.48±1.2          & 94.07±1.5          & \textbf{63.49±0.4} & 66.89±2.8          & 93.73±0.9          & 72.69±1.9          & \textbf{91.92±0.6} \\
\multicolumn{1}{c|}{}                                                         & \multicolumn{1}{c|}{}                                                 & \multicolumn{1}{c|}{$top_k=10$}       & \textbf{95.21±0.5} & \textbf{94.56±1.1} & 63.33±0.4          & \textbf{69.91±1.1} & 94.67±0.1          & 71.84±1.5          & 90.05±1.2          \\
\multicolumn{1}{c|}{}                                                         & \multicolumn{1}{c|}{}                                                 & \multicolumn{1}{c|}{HDRF}             & 94.02±0.9          & 94.91±0.1          & 64.78±1.6          & 70.65±2.6          & 95.01±0.1          & OOM                & OOM                \\
\multicolumn{1}{c|}{}                                                         & \multicolumn{1}{c|}{}                                                 & \multicolumn{1}{c|}{w/o Partitioning} & 94.62±0.5          & 97.11±0.3          & 76.50±1.8          & 68.77±3.0          & N/A                & OOM                & OOM                \\ \cmidrule{2-10} 
\multicolumn{1}{c|}{}                                                         & \multicolumn{1}{c|}{\multirow{6}{*}{\rotatebox[origin=c]{90}{DyRep}}} & \multicolumn{1}{c|}{$top_k=0$}        & 90.68±0.9          & 93.07±0.4          & 59.69±3.2          & 58.66±1.9          & \textbf{93.99±1.4} & \textbf{78.70±0.2} & 68.71±12.2         \\
\multicolumn{1}{c|}{}                                                         & \multicolumn{1}{c|}{}                                                 & \multicolumn{1}{c|}{$top_k=1$}        & 81.82±1.7          & 87.14±3.8          & 60.15±1.3          & 55.77±3.8          & 87.38±4.6          & 77.86±0.1          & 87.07±1.9          \\
\multicolumn{1}{c|}{}                                                         & \multicolumn{1}{c|}{}                                                 & \multicolumn{1}{c|}{$top_k=5$}        & 90.29±1.3          & \textbf{93.47±1.2} & \textbf{62.76±6.6} & 53.53±2.1          & 92.74±0.7          & 77.73±0.1          & \textbf{89.45±0.5} \\
\multicolumn{1}{c|}{}                                                         & \multicolumn{1}{c|}{}                                                 & \multicolumn{1}{c|}{$top_k=10$}       & \textbf{91.58±1.5} & 92.24±0.3          & 60.27±8.4          & \textbf{61.56±2.1} & 92.01±1.4          & 78.15±0.1          & 83.31±3.8          \\
\multicolumn{1}{c|}{}                                                         & \multicolumn{1}{c|}{}                                                 & \multicolumn{1}{c|}{HDRF}             & 91.21±0.7          & \textbf{93.61±1.0} & 59.16±1.5          & 71.43±2.2          & 93.76±0.5          & OOM                & OOM                \\
\multicolumn{1}{c|}{}                                                         & \multicolumn{1}{c|}{}                                                 & \multicolumn{1}{c|}{w/o Partitioning} & 94.59±0.2          & 97.98±0.1          & 75.37±1.7          & 68.77±2.1          & N/A                & OOM                & OOM                \\ \cmidrule{2-10} 
\multicolumn{1}{c|}{}                                                         & \multicolumn{1}{c|}{\multirow{6}{*}{\rotatebox[origin=c]{90}{TGN}}}   & \multicolumn{1}{c|}{$top_k=0$}        & 97.50±0.2          & \textbf{96.69±0.3} & 61.54±0.8          & 55.97±4.3          & 93.39±1.0          & 82.16±0.3          & 85.75±1.3          \\
\multicolumn{1}{c|}{}                                                         & \multicolumn{1}{c|}{}                                                 & \multicolumn{1}{c|}{$top_k=1$}        & 95.90±0.1          & 92.50±1.1          & 77.34±5.4          & \textbf{62.79±6.0} & 91.32±5.5          & \textbf{82.27±0.2} & 84.22±2.9          \\
\multicolumn{1}{c|}{}                                                         & \multicolumn{1}{c|}{}                                                 & \multicolumn{1}{c|}{$top_k=5$}        & 97.35±0.1          & 95.84±1.2          & 81.44±2.7          & 58.98±3.7          & 94.20±0.7          & 82.08±0.2          & 81.93±8.6          \\
\multicolumn{1}{c|}{}                                                         & \multicolumn{1}{c|}{}                                                 & \multicolumn{1}{c|}{$top_k=10$}       & \textbf{97.61±0.1} & 95.53±1.4          & \textbf{85.66±0.9} & 53.64±5.4          & \textbf{94.51±0.5} & 81.98±0.1          & \textbf{85.93±2.0} \\
\multicolumn{1}{c|}{}                                                         & \multicolumn{1}{c|}{}                                                 & \multicolumn{1}{c|}{HDRF}             & 96.36±1.9          & 94.59±2.0          & 78.37±8.4          & 53.47±3.6          & 94.53±0.2          & OOM                & OOM                \\
\multicolumn{1}{c|}{}                                                         & \multicolumn{1}{c|}{}                                                 & \multicolumn{1}{c|}{w/o Partitioning} & 98.46±0.1          & 98.70±0.1          & 85.88±3.0          & 71.76±5.3          & N/A                & OOM                & OOM                \\ \cmidrule{2-10} 
\multicolumn{1}{c|}{}                                                         & \multicolumn{1}{c|}{\multirow{6}{*}{\rotatebox[origin=c]{90}{TIGE}}}  & \multicolumn{1}{c|}{$top_k=0$}        & 98.34±0.0          & 98.19±0.1          & 75.32±1.1          & \textbf{83.05±0.7} & 93.09±2.4          & 82.11±0.1          & 88.03±7.4          \\
\multicolumn{1}{c|}{}                                                         & \multicolumn{1}{c|}{}                                                 & \multicolumn{1}{c|}{$top_k=1$}        & 98.18±0.1          & 97.81±0.2          & 84.80±1.0          & 82.20±0.7          & 92.32±2.3          & 82.41±0.1          & 88.81±3.0          \\
\multicolumn{1}{c|}{}                                                         & \multicolumn{1}{c|}{}                                                 & \multicolumn{1}{c|}{$top_k=5$}        & 98.22±0.1          & \textbf{98.29±0.4} & 86.80±1.9          & 82.43±0.6          & 92.73±2.8          & \textbf{82.75±0.1} & \textbf{89.88±4.2} \\
\multicolumn{1}{c|}{}                                                         & \multicolumn{1}{c|}{}                                                 & \multicolumn{1}{c|}{$top_k=10$}       & \textbf{98.50±0.0} & 98.07±0.1          & \textbf{88.28±1.6} & 82.59±0.3          & \textbf{94.12±0.5} & 82.57±0.2          & 89.04±2.9          \\
\multicolumn{1}{c|}{}                                                         & \multicolumn{1}{c|}{}                                                 & \multicolumn{1}{c|}{HDRF}             & 97.98±0.4          & 97.89±0.3          & 86.17±0.6          & 82.78±0.8          & 91.98±0.0          & OOM                & OOM                \\
\multicolumn{1}{c|}{}                                                         & \multicolumn{1}{c|}{}                                                 & \multicolumn{1}{c|}{w/o Partitioning} & 98.83±0.1          & 99.04±0.0          & 89.64±0.9          & 87.85±0.9          & N/A                & OOM                & OOM                \\ \midrule
\multicolumn{1}{c|}{\multirow{24}{*}{\rotatebox[origin=c]{90}{Inductive}}}    & \multicolumn{1}{c|}{\multirow{6}{*}{\rotatebox[origin=c]{90}{Jodie}}} & \multicolumn{1}{c|}{$top_k=0$}        & 93.59±0.7          & 92.86±0.2          & 61.71±0.3          & 70.43±5.6          & 91.75±2.2          & 68.76±1.6          & 79.70±2.1          \\
\multicolumn{1}{c|}{}                                                         & \multicolumn{1}{c|}{}                                                 & \multicolumn{1}{c|}{$top_k=1$}        & 88.95±0.7          & 91.39±1.2          & 60.37±1.3          & 74.58±0.7          & 93.21±0.6          & 68.91±0.1          & 79.71±0.7          \\
\multicolumn{1}{c|}{}                                                         & \multicolumn{1}{c|}{}                                                 & \multicolumn{1}{c|}{$top_k=5$}        & 92.99±1.4          & 93.18±1.7          & 61.82±0.5          & 74.53±3.3          & 93.31±1.0          & 69.21±0.5          & \textbf{81.18±1.0} \\
\multicolumn{1}{c|}{}                                                         & \multicolumn{1}{c|}{}                                                 & \multicolumn{1}{c|}{$top_k=10$}       & \textbf{93.96±0.5} & \textbf{93.46±1.1} & \textbf{62.98±1.1} & \textbf{79.68±1.2} & \textbf{94.27±0.1} & \textbf{69.59±0.2} & 78.25±2.7          \\
\multicolumn{1}{c|}{}                                                         & \multicolumn{1}{c|}{}                                                 & \multicolumn{1}{c|}{HDRF}             & 92.29±0.6          & 93.05±0.6          & 64.90±0.9          & 82.67±0.9          & 94.66±0.1          & OOM                & OOM                \\
\multicolumn{1}{c|}{}                                                         & \multicolumn{1}{c|}{}                                                 & \multicolumn{1}{c|}{w/o Partitioning} & 93.11±0.4          & 94.36±1.1          & 77.83±2.1          & 82.55±1.9          & N/A                & OOM                & OOM                \\ \cmidrule{2-10} 
\multicolumn{1}{c|}{}                                                         & \multicolumn{1}{c|}{\multirow{6}{*}{\rotatebox[origin=c]{90}{DyRep}}} & \multicolumn{1}{c|}{$top_k=0$}        & 89.40±0.5          & 93.11±0.4          & 60.05±4.1          & 69.82±1.6          & \textbf{93.68±1.3} & \textbf{64.28±0.1} & 59.07±6.8          \\
\multicolumn{1}{c|}{}                                                         & \multicolumn{1}{c|}{}                                                 & \multicolumn{1}{c|}{$top_k=1$}        & 81.33±1.4          & 90.00±2.9          & 58.68±1.5          & 63.70±6.7          & 86.72±4.9          & 64.18±0.3          & 75.25±2.8          \\
\multicolumn{1}{c|}{}                                                         & \multicolumn{1}{c|}{}                                                 & \multicolumn{1}{c|}{$top_k=5$}        & 89.12±1.4          & \textbf{93.78±0.8} & \textbf{62.82±6.1} & 61.03±3.1          & 92.43±0.7          & 64.20±0.3          & \textbf{78.35±0.8} \\
\multicolumn{1}{c|}{}                                                         & \multicolumn{1}{c|}{}                                                 & \multicolumn{1}{c|}{$top_k=10$}       & \textbf{90.14±1.2} & 92.11±0.5          & 59.57±7.4          & \textbf{74.42±2.9} & 91.63±1.5          & 64.12±0.2          & 71.27±4.7          \\
\multicolumn{1}{c|}{}                                                         & \multicolumn{1}{c|}{}                                                 & \multicolumn{1}{c|}{HDRF}             & 89.70±0.2          & 92.62±0.9          & 59.45±0.5          & 82.67±0.9          & 93.66±0.4          & OOM                & OOM                \\
\multicolumn{1}{c|}{}                                                         & \multicolumn{1}{c|}{}                                                 & \multicolumn{1}{c|}{w/o Partitioning} & 92.05±0.3          & 95.68±0.2          & 78.55±1.1          & 81.33±2.1          & N/A                & OOM                & OOM                \\ \cmidrule{2-10} 
\multicolumn{1}{c|}{}                                                         & \multicolumn{1}{c|}{\multirow{6}{*}{\rotatebox[origin=c]{90}{TGN}}}   & \multicolumn{1}{c|}{$top_k=0$}        & 96.86±0.2          & \textbf{94.84±0.5} & 76.14±1.0          & 59.32±3.8          & 93.29±0.9          & 66.11±0.7          & \textbf{74.08±0.2} \\
\multicolumn{1}{c|}{}                                                         & \multicolumn{1}{c|}{}                                                 & \multicolumn{1}{c|}{$top_k=1$}        & 95.44±0.4          & 90.71±0.3          & 77.53±3.5          & \textbf{66.94±8.4} & 91.09±5.5          & 66.51±0.7          & 68.58±2.3          \\
\multicolumn{1}{c|}{}                                                         & \multicolumn{1}{c|}{}                                                 & \multicolumn{1}{c|}{$top_k=5$}        & 96.62±0.1          & 93.95±1.7          & 81.20±3.2          & 63.79±5.9          & 93.80±0.8          & 66.82±0.4          & 70.23±6.0          \\
\multicolumn{1}{c|}{}                                                         & \multicolumn{1}{c|}{}                                                 & \multicolumn{1}{c|}{$top_k=10$}       & \textbf{96.90±0.1} & 93.49±1.9          & \textbf{86.29±0.7} & 55.40±8.3          & \textbf{94.10±0.5} & \textbf{67.39±0.8} & 73.34±2.6          \\
\multicolumn{1}{c|}{}                                                         & \multicolumn{1}{c|}{}                                                 & \multicolumn{1}{c|}{HDRF}             & 96.05±1.4          & 92.31±2.6          & 79.47±8.3          & 52.91±10.8         & 94.07±0.2          & OOM                & OOM                \\
\multicolumn{1}{c|}{}                                                         & \multicolumn{1}{c|}{}                                                 & \multicolumn{1}{c|}{w/o Partitioning} & 97.81±0.1          & 97.55±0.1          & 85.55±2.9          & 80.42±4.9          & N/A                & OOM                & OOM                \\ \cmidrule{2-10} 
\multicolumn{1}{c|}{}                                                         & \multicolumn{1}{c|}{\multirow{6}{*}{\rotatebox[origin=c]{90}{TIGE}}}  & \multicolumn{1}{c|}{$top_k=0$}        & 98.05±0.1          & 97.40±0.3          & 79.30±3.6          & \textbf{85.32±1.1} & 92.59±2.6          & \textbf{65.98±0.4} & 79.43±4.9          \\
\multicolumn{1}{c|}{}                                                         & \multicolumn{1}{c|}{}                                                 & \multicolumn{1}{c|}{$top_k=1$}        & 97.87±0.1          & 96.86±0.3          & 84.41±1.7          & 85.27±0.9          & 92.27±2.1          & 65.41±0.2          & 76.78±4.5          \\
\multicolumn{1}{c|}{}                                                         & \multicolumn{1}{c|}{}                                                 & \multicolumn{1}{c|}{$top_k=5$}        & 97.91±0.1          & \textbf{97.50±0.4} & 86.30±1.8          & 84.84±1.1          & 92.31±3.0          & 65.98±0.7          & \textbf{79.46±2.6} \\
\multicolumn{1}{c|}{}                                                         & \multicolumn{1}{c|}{}                                                 & \multicolumn{1}{c|}{$top_k=10$}       & \textbf{98.17±0.0} & 97.25±0.2          & \textbf{87.90±1.6} & 85.12±0.4          & \textbf{93.79±0.5} & 65.60±0.3          & 78.37±2.3          \\
\multicolumn{1}{c|}{}                                                         & \multicolumn{1}{c|}{}                                                 & \multicolumn{1}{c|}{HDRF}             & 97.68±0.2          & 96.96±0.2          & 86.38±0.6          & 84.44±0.6          & 91.62±0.1          & OOM                & OOM                \\
\multicolumn{1}{c|}{}                                                         & \multicolumn{1}{c|}{}                                                 & \multicolumn{1}{c|}{w/o Partitioning} & 98.45±0.1          & 98.39±0.1          & 89.51±0.7          & 90.14±1.0          & N/A                & OOM                & OOM                \\ \bottomrule
\end{tabular}

\label{tab:performance}
\end{table*}

%% file: tab_nodes.tex
\useunder{\uline}{\ul}{}

\begin{table}[t]
\centering
\caption{AUROC (\%) for dynamic/static node classification task. We use the results reported in \cite{tiger} which trained without graph partitioning as baselines and present the results of HDRF algorithm for comparing. The best AUROC of different $top_k$ on different datasets and backbones are in bold.}
\begin{tabular}{ccccc}
\toprule
                                                                              &                                       & Wikipeida          & Reddit             & MOOC               \\ \midrule
\multicolumn{1}{c|}{\multirow{6}{*}{\rotatebox[origin=c]{90}{Jodie}}}         & \multicolumn{1}{c|}{$top_k=0$}        & 87.54±0.7          & 58.98±3.6          & 64.27±1.2          \\
\multicolumn{1}{c|}{}                                                         & \multicolumn{1}{c|}{$top_k=1$}        & \textbf{87.83±0.2} & 62.04±2.9          & 62.28±0.9          \\
\multicolumn{1}{c|}{}                                                         & \multicolumn{1}{c|}{$top_k=5$}        & 87.52±1.1          & \textbf{64.00±5.5} & 61.11±0.7          \\
\multicolumn{1}{c|}{}                                                         & \multicolumn{1}{c|}{$top_k=10$}       & 86.97±1.0          & 61.90±0.4          & \textbf{65.36±1.8} \\
\multicolumn{1}{c|}{}                                                         & \multicolumn{1}{c|}{HDRF}             & 87.82±0.5          & 63.26±5.5          & 65.72±1.6          \\
\multicolumn{1}{c|}{}                                                         & \multicolumn{1}{c|}{w/o Partitioning} & 84.84±1.2          & 61.83±2.7          & 66.87±0.4          \\ \midrule
\multicolumn{1}{c|}{\multirow{6}{*}{\rotatebox[origin=c]{90}{DyRep}}}         & \multicolumn{1}{c|}{$top_k=0$}        & 85.92±0.8          & \textbf{64.27±1.4} & 63.58±1.6          \\
\multicolumn{1}{c|}{}                                                         & \multicolumn{1}{c|}{$top_k=1$}        & 86.53±0.9          & 62.29±5.1          & 62.12±3.1          \\
\multicolumn{1}{c|}{}                                                         & \multicolumn{1}{c|}{$top_k=5$}        & \textbf{86.63±2.7} & 62.01±1.4          & 64.12±1.5          \\
\multicolumn{1}{c|}{}                                                         & \multicolumn{1}{c|}{$top_k=10$}       & 86.26±1.2          & 61.88±4.4          & \textbf{65.43±3.1} \\
\multicolumn{1}{c|}{}                                                         & \multicolumn{1}{c|}{HDRF}             & 86.95±0.4          & 63.59±1.3          & 63.90±3.8          \\
\multicolumn{1}{c|}{}                                                         & \multicolumn{1}{c|}{w/o Partitioning} & 84.59±2.2          & 62.91±2.4          & 67.76±0.5          \\ \midrule
\multicolumn{1}{c|}{\multirow{6}{*}{\rotatebox[origin=c]{90}{TGN}}}           & \multicolumn{1}{c|}{$top_k=0$}        & \textbf{86.39±1.0} & \textbf{67.77±4.5} & \textbf{73.61±1.1} \\
\multicolumn{1}{c|}{}                                                         & \multicolumn{1}{c|}{$top_k=1$}        & 83.71±2.5          & 62.81±1.3          & 72.04±0.7          \\
\multicolumn{1}{c|}{}                                                         & \multicolumn{1}{c|}{$top_k=5$}        & 82.40±1.8          & {\ul 67.31±1.1}    & {\ul 72.95±1.3}    \\
\multicolumn{1}{c|}{}                                                         & \multicolumn{1}{c|}{$top_k=10$}       & 82.64±2.0          & 66.88±3.2          & 71.81±0.8          \\
\multicolumn{1}{c|}{}                                                         & \multicolumn{1}{c|}{HDRF}             & 83.93±1.2          & 64.41±2.0          & 72.40±1.1          \\
\multicolumn{1}{c|}{}                                                         & \multicolumn{1}{c|}{w/o Partitioning} & 87.81±0.3          & 67.06±0.9          & 59.54±1.0          \\ \midrule
\multicolumn{1}{c|}{\multirow{6}{*}{\textbf{\rotatebox[origin=c]{90}{TIGE}}}} & \multicolumn{1}{c|}{$top_k=0$}        & 85.59±3.2          & \textbf{64.16±0.5} & 73.08±1.8          \\
\multicolumn{1}{c|}{}                                                         & \multicolumn{1}{c|}{$top_k=1$}        & \textbf{86.89±1.4} & 63.59±2.4          & 72.50±0.5          \\
\multicolumn{1}{c|}{}                                                         & \multicolumn{1}{c|}{$top_k=5$}        & 85.32±0.8          & 63.44±3.6          & \textbf{73.63±0.7} \\
\multicolumn{1}{c|}{}                                                         & \multicolumn{1}{c|}{$top_k=10$}       & 85.01±0.4          & 62.23±0.6          & 72.86±1.1          \\
\multicolumn{1}{c|}{}                                                         & \multicolumn{1}{c|}{HDRF}             & 85.91±2.5          & 63.74±1.1          & 72.69±0.9          \\
\multicolumn{1}{c|}{}                                                         & \multicolumn{1}{c|}{w/o Partitioning} & 86.92±0.7          & 69.41±1.3          & 72.35±2.3          \\ \bottomrule
\end{tabular}
\label{tab:node_class}
\end{table}

%% file: tab_cut.tex
\begin{table}[t]
\centering
\caption{Edge cut\%, average nodes portion and standard deviation of edges and nodes on each GPU, i.e., partitions, of dataset Taobao. Note that the Taobao has 1e8 edges and 5e7 nodes.}
\begin{tabular}{cc|cc|cc}
\toprule
\multicolumn{2}{c|}{\multirow{2}{*}{Dataset: Taobao}}                             & \multicolumn{2}{c|}{Edge Statistics} & \multicolumn{2}{c}{Node Statistics} \\ \cmidrule{3-6} 
\multicolumn{2}{c|}{}                                                             & Total Cut           & Std.           & Avg. Portion         & Std.         \\ \midrule
\multicolumn{1}{c|}{}                                                & KL         & 20.5\%              & 3.2e7          & 25.0\%               & 0.5          \\ \midrule
\multicolumn{1}{c|}{\multirow{4}{*}{\rotatebox[origin=c]{90}{Ours}}} & $top_k=0$  & 69.5\%              & 3.1e3          & 25.0\%               & 5.9e3        \\
\multicolumn{1}{c|}{}                                                & $top_k=1$  & 40.1\%              & 3.1e3          & 25.5\%               & 1.1e4        \\
\multicolumn{1}{c|}{}                                                & $top_k=5$  & 21.4\%              & 4.1e2          & 28.7\%               & 4.9e3        \\
\multicolumn{1}{c|}{}                                                & $top_k=10$ & 8.5\%               & 1.9e2          & 32.4\%               & 8.8e3        \\ \midrule
\multicolumn{1}{c|}{}                                                & HDRF       & 0\%                 & 2.2            & 50.5\%               & 2.1e3        \\
\multicolumn{1}{c|}{}                                                & Random     & 75.1\%              & 8.6e5          & 25.0\%               & 1.2e3        \\ \bottomrule
\end{tabular}
\label{tab:cut}
\end{table}

%% file: tab_kl.tex
\begin{table*}[!thbp]
\centering
\caption{Link prediction task performance results are in Average Precision (\%). Training times are in seconds per epoch. $top_k=0$ refers to our methods. Note that our methods with other $top_k$ value may outperform KL on downstream tasks.}

\begin{tabular}{ccccccccccc}
\toprule
                                                                          &                            & \multicolumn{3}{c}{ML25m}                                                                                                         & \multicolumn{3}{c}{DGraphFin}                                                                                                     & \multicolumn{3}{c}{Taobao}                                                                                   \\ \midrule
                                                                          &                            & Transductive       & Inductive          & \multicolumn{1}{c|}{\begin{tabular}[c]{@{}c@{}}Training Time\\ (Speed-up)\end{tabular}} & Transductive       & Inductive          & \multicolumn{1}{c|}{\begin{tabular}[c]{@{}c@{}}Training Time\\ (Speed-up)\end{tabular}} & Transductive       & Inductive          & \begin{tabular}[c]{@{}c@{}}Training Time\\ (Speed-up)\end{tabular} \\ \midrule
\multicolumn{1}{c|}{\multirow{4}{*}{\rotatebox[origin=c]{90}{KL}}}        & \multicolumn{1}{c|}{Jodie} & \textbf{93.01±0.6} & \textbf{93.31±0.4} & \multicolumn{1}{c|}{1658.6}                                                             & 74.53±0.7          & \textbf{68.97±0.7} & \multicolumn{1}{c|}{115.5}                                                              & 89.73±4.0          & 78.22±5.1          & 9602.8                                                             \\
\multicolumn{1}{c|}{}                                                     & \multicolumn{1}{c|}{DyRep} & 92.10±2.5          & 92.49±1.8          & \multicolumn{1}{c|}{1723.9}                                                             & 78.56±0.1          & \textbf{64.98±0.2} & \multicolumn{1}{c|}{118.8}                                                              & \textbf{77.97±6.7} & \textbf{68.30±5.8} & 10347.0                                                            \\
\multicolumn{1}{c|}{}                                                     & \multicolumn{1}{c|}{TGN}   & 91.58±1.7          & 92.06±1.4          & \multicolumn{1}{c|}{1721.0}                                                             & \textbf{82.74±0.4} & \textbf{66.68±1.2} & \multicolumn{1}{c|}{122.0}                                                              & 83.15±3.9          & 71.29±1.2          & 10466.8                                                            \\
\multicolumn{1}{c|}{}                                                     & \multicolumn{1}{c|}{TIGE}  & 88.69±3.1          & 88.67±3.3          & \multicolumn{1}{c|}{1732.8}                                                             & \textbf{82.74±0.1} & \textbf{66.23±0.2} & \multicolumn{1}{c|}{123.1}                                                              & \textbf{89.62±3.8} & \textbf{80.21±2.6} & 10492.7                                                            \\ \midrule
\multicolumn{1}{c|}{\multirow{4}{*}{\rotatebox[origin=c]{90}{$top_k=0$}}} & \multicolumn{1}{c|}{Jodie} & 92.22±2.1          & 91.75±2.2          & \multicolumn{1}{c|}{\textbf{226.5 (7.3x)}}                                              & \textbf{74.75±1.7} & 68.76±1.6          & \multicolumn{1}{c|}{\textbf{102.7 (1.1x)}}                                              & \textbf{91.43±1.1} & \textbf{79.70±2.1} & \textbf{917.3 (10.5x)}                                             \\
\multicolumn{1}{c|}{}                                                     & \multicolumn{1}{c|}{DyRep} & \textbf{93.99±1.4} & \textbf{93.68±1.3} & \multicolumn{1}{c|}{\textbf{237.0 (7.3x)}}                                              & \textbf{78.70±0.2} & 64.28±0.1          & \multicolumn{1}{c|}{\textbf{105.2 (1.1x)}}                                              & 68.71±12.2         & 59.07±6.8          & \textbf{961.8 (10.7x)}                                             \\
\multicolumn{1}{c|}{}                                                     & \multicolumn{1}{c|}{TGN}   & \textbf{93.39±1.0} & \textbf{93.29±0.9} & \multicolumn{1}{c|}{\textbf{238.4 (7.2x)}}                                              & 82.16±0.3          & 66.11±0.7          & \multicolumn{1}{c|}{\textbf{107.5 (1.1x)}}                                              & \textbf{85.75±1.3} & \textbf{74.08±0.2} & \textbf{976.6 (10.7x)}                                             \\
\multicolumn{1}{c|}{}                                                     & \multicolumn{1}{c|}{TIGE}  & \textbf{93.09±2.4} & \textbf{92.59±2.6} & \multicolumn{1}{c|}{\textbf{251.4 (6.9x)}}                                              & 82.11±0.1          & 65.98±0.4          & \multicolumn{1}{c|}{\textbf{108.1 (1.1x)}}                                              & 88.03±7.4          & 79.43±4.9          & \textbf{984.4 (10.7x)}                                             \\ \bottomrule
\end{tabular}

\label{tab:kl}
\end{table*}

%% file: tab_par_speedup.tex
\begin{table}[t]
\centering
\caption{Partitioning time (in seconds) comparison between \oursp~and KL for four datasets. Experiments are performed on CPU. }
\begin{tabular}{ccccc}
\toprule  & Wikipedia     &                   DGraphFin    &                   ML25m   &                   Taobao     \\ \midrule &
 \multicolumn{1}{c|}{\begin{tabular}[c]{@{}c@{}}Time\\(Speed-up)\end{tabular} }&  \multicolumn{1}{c|}{\begin{tabular}[c]{@{}c@{}}Time\\(Speed-up)\end{tabular} }  &  \multicolumn{1}{c|}{\begin{tabular}[c]{@{}c@{}}Time\\(Speed-up)\end{tabular}} & \begin{tabular}[c]{@{}c@{}}Time\\(Speed-up)\end{tabular} \\ \midrule
                                                                      \multicolumn{1}{c|}{KL}  &                                 \multicolumn{1}{c|}{11.01}                                              &                                                \multicolumn{1}{c|}{238.46}  & \multicolumn{1}{c|}{866.85 }& 14862.33                                                \\
         \multicolumn{1}{c|}{\oursp}  &                                              \multicolumn{1}{c|}{ \textbf{0.27 (41x)}  } &
 \multicolumn{1}{c|}{\textbf{3.12 (76.5x)} }   & \multicolumn{1}{c|}{\textbf{10.67 (81.24x)} }  &      \textbf{157.18 (94.57x)}                                                                      \\
\bottomrule
\end{tabular}

\label{tab:par_speedup}
\end{table}

%% file: main.bbl
\begin{thebibliography}{10}

\bibitem{jodie}
Srijan Kumar, Xikun Zhang, and Jure Leskovec.
\newblock Predicting dynamic embedding trajectory in temporal interaction
  networks.
\newblock In {\em Proceedings of the 25th ACM SIGKDD international conference
  on knowledge discovery \& data mining}, pages 1269--1278, 2019.

\bibitem{dyrep}
Rakshit Trivedi, Mehrdad Farajtabar, Prasenjeet Biswal, and Hongyuan Zha.
\newblock Dyrep: Learning representations over dynamic graphs.
\newblock In {\em International conference on learning representations}, 2019.

\bibitem{tgat}
Da~Xu, Chuanwei Ruan, Evren Korpeoglu, Sushant Kumar, and Kannan Achan.
\newblock Inductive representation learning on temporal graphs.
\newblock {\em arXiv preprint arXiv:2002.07962}, 2020.

\bibitem{tgn}
Emanuele Rossi, Ben Chamberlain, Fabrizio Frasca, Davide Eynard, Federico
  Monti, and Michael Bronstein.
\newblock Temporal graph networks for deep learning on dynamic graphs.
\newblock {\em arXiv preprint arXiv:2006.10637}, 2020.

\bibitem{tiger}
Yao Zhang, Yun Xiong, Yongxiang Liao, Yiheng Sun, Yucheng Jin, Xuehao Zheng,
  and Yangyong Zhu.
\newblock Tiger: Temporal interaction graph embedding with restarts.
\newblock In {\em Proceedings of the ACM Web Conference 2023}, pages 478--488,
  2023.

\bibitem{edge}
Xinshi Chen, Yan Zhu, Haowen Xu, Mengyang Liu, Liang Xiong, Muhan Zhang, and
  Le~Song.
\newblock Efficient dynamic graph representation learning at scale.
\newblock {\em arXiv preprint arXiv:2112.07768}, 2021.

\bibitem{metis}
George Karypis and Vipin Kumar.
\newblock A fast and high quality multilevel scheme for partitioning irregular
  graphs.
\newblock {\em SIAM Journal on scientific Computing}, 20(1):359--392, 1998.

\bibitem{kl}
Brian~W Kernighan and Shen Lin.
\newblock An efficient heuristic procedure for partitioning graphs.
\newblock {\em The Bell system technical journal}, 49(2):291--307, 1970.

\bibitem{euler}
Euler github.
\newblock \url{https://github.com/alibaba/euler}.

\bibitem{lgd}
Isabelle Stanton and Gabriel Kliot.
\newblock Streaming graph partitioning for large distributed graphs.
\newblock In {\em Proceedings of the 18th ACM SIGKDD international conference
  on Knowledge discovery and data mining}, pages 1222--1230, 2012.

\bibitem{roc}
Zhihao Jia, Sina Lin, Mingyu Gao, Matei Zaharia, and Alex Aiken.
\newblock Improving the accuracy, scalability, and performance of graph neural
  networks with roc.
\newblock {\em Proceedings of Machine Learning and Systems}, 2:187--198, 2020.

\bibitem{libra}
Cong Xie, Ling Yan, Wu-Jun Li, and Zhihua Zhang.
\newblock Distributed power-law graph computing: Theoretical and empirical
  analysis.
\newblock {\em Advances in neural information processing systems}, 27, 2014.

\bibitem{greedy}
Joseph~E Gonzalez, Yucheng Low, Haijie Gu, Danny Bickson, and Carlos Guestrin.
\newblock Powergraph: Distributed graph-parallel computation on natural graphs.
\newblock In {\em Presented as part of the 10th $\{$USENIX$\}$ Symposium on
  Operating Systems Design and Implementation ($\{$OSDI$\}$ 12)}, pages 17--30,
  2012.

\bibitem{hdrf}
Fabio Petroni, Leonardo Querzoni, Khuzaima Daudjee, Shahin Kamali, and Giorgio
  Iacoboni.
\newblock Hdrf: Stream-based partitioning for power-law graphs.
\newblock In {\em Proceedings of the 24th ACM international on conference on
  information and knowledge management}, pages 243--252, 2015.

\bibitem{pint}
Amauri Souza, Diego Mesquita, Samuel Kaski, and Vikas Garg.
\newblock Provably expressive temporal graph networks.
\newblock {\em Advances in Neural Information Processing Systems},
  35:32257--32269, 2022.

\bibitem{cawn}
Yanbang Wang, Yen-Yu Chang, Yunyu Liu, Jure Leskovec, and Pan Li.
\newblock Inductive representation learning in temporal networks via causal
  anonymous walks.
\newblock {\em arXiv preprint arXiv:2101.05974}, 2021.

\bibitem{zebra}
Yiming Li, Yanyan Shen, Lei Chen, and Mingxuan Yuan.
\newblock Zebra: When temporal graph neural networks meet temporal personalized
  pagerank.
\newblock {\em Proceedings of the VLDB Endowment}, 16(6):1332--1345, 2023.

\bibitem{attack}
Reuven Cohen, Keren Erez, Daniel Ben-Avraham, and Shlomo Havlin.
\newblock Breakdown of the internet under intentional attack.
\newblock {\em Physical review letters}, 86(16):3682, 2001.

\bibitem{ml25m}
F~Maxwell Harper and Joseph~A Konstan.
\newblock The movielens datasets: History and context.
\newblock {\em Acm transactions on interactive intelligent systems (tiis)},
  5(4):1--19, 2015.

\bibitem{dgraph}
Xuanwen Huang, Yang Yang, Yang Wang, Chunping Wang, Zhisheng Zhang, Jiarong Xu,
  and Lei Chen.
\newblock {DG}raph: A large-scale financial dataset for graph anomaly
  detection.
\newblock In {\em Thirty-sixth Conference on Neural Information Processing
  Systems Datasets and Benchmarks Track}, 2022.

\bibitem{taobao}
Jingwei Zhuo, Ziru Xu, Wei Dai, Han Zhu, Han Li, Jian Xu, and Kun Gai.
\newblock Learning optimal tree models under beam search.
\newblock In {\em International Conference on Machine Learning}, pages
  11650--11659. PMLR, 2020.

\bibitem{sage}
Will Hamilton, Zhitao Ying, and Jure Leskovec.
\newblock Inductive representation learning on large graphs.
\newblock {\em Advances in neural information processing systems}, 30, 2017.

\bibitem{vrgcn}
Jianfei Chen, Jun Zhu, and Le~Song.
\newblock Stochastic training of graph convolutional networks with variance
  reduction.
\newblock {\em arXiv preprint arXiv:1710.10568}, 2017.

\bibitem{fastgcn}
Jie Chen, Tengfei Ma, and Cao Xiao.
\newblock Fastgcn: fast learning with graph convolutional networks via
  importance sampling.
\newblock {\em arXiv preprint arXiv:1801.10247}, 2018.

\bibitem{clustergcn}
Wei-Lin Chiang, Xuanqing Liu, Si~Si, Yang Li, Samy Bengio, and Cho-Jui Hsieh.
\newblock Cluster-gcn: An efficient algorithm for training deep and large graph
  convolutional networks.
\newblock In {\em Proceedings of the 25th ACM SIGKDD international conference
  on knowledge discovery \& data mining}, pages 257--266, 2019.

\bibitem{neugraph}
Lingxiao Ma, Zhi Yang, Youshan Miao, Jilong Xue, Ming Wu, Lidong Zhou, and
  Yafei Dai.
\newblock Neugraph: Parallel deep neural network computation on large graphs.
\newblock In {\em USENIX Annual Technical Conference}, pages 443--458, 2019.

\bibitem{dgcl}
Zhenkun Cai, Xiao Yan, Yidi Wu, Kaihao Ma, James Cheng, and Fan Yu.
\newblock Dgcl: an efficient communication library for distributed gnn
  training.
\newblock In {\em Proceedings of the Sixteenth European Conference on Computer
  Systems}, pages 130--144, 2021.

\bibitem{distgnn}
Vasimuddin Md, Sanchit Misra, Guixiang Ma, Ramanarayan Mohanty, Evangelos
  Georganas, Alexander Heinecke, Dhiraj Kalamkar, Nesreen~K Ahmed, and
  Sasikanth Avancha.
\newblock Distgnn: Scalable distributed training for large-scale graph neural
  networks.
\newblock In {\em Proceedings of the International Conference for High
  Performance Computing, Networking, Storage and Analysis}, pages 1--14, 2021.

\bibitem{aligraph}
Rong Zhu, Kun Zhao, Hongxia Yang, Wei Lin, Chang Zhou, Baole Ai, Yong Li, and
  Jingren Zhou.
\newblock Aligraph: A comprehensive graph neural network platform.
\newblock {\em arXiv preprint arXiv:1902.08730}, 2019.

\bibitem{agl}
Dalong Zhang, Xin Huang, Ziqi Liu, Zhiyang Hu, Xianzheng Song, Zhibang Ge,
  Zhiqiang Zhang, Lin Wang, Jun Zhou, Yang Shuang, et~al.
\newblock Agl: a scalable system for industrial-purpose graph machine learning.
\newblock {\em arXiv preprint arXiv:2003.02454}, 2020.

\bibitem{distdgl}
Da~Zheng, Chao Ma, Minjie Wang, Jinjing Zhou, Qidong Su, Xiang Song, Quan Gan,
  Zheng Zhang, and George Karypis.
\newblock Distdgl: distributed graph neural network training for billion-scale
  graphs.
\newblock In {\em 2020 IEEE/ACM 10th Workshop on Irregular Applications:
  Architectures and Algorithms (IA3)}, pages 36--44. IEEE, 2020.

\bibitem{bgl}
Tianfeng Liu, Yangrui Chen, Dan Li, Chuan Wu, Yibo Zhu, Jun He, Yanghua Peng,
  Hongzheng Chen, Hongzhi Chen, and Chuanxiong Guo.
\newblock $\{$BGL$\}$:$\{$GPU-Efficient$\}$$\{$GNN$\}$ training by optimizing
  graph data $\{$I/O$\}$ and preprocessing.
\newblock In {\em 20th USENIX Symposium on Networked Systems Design and
  Implementation (NSDI 23)}, pages 103--118, 2023.

\bibitem{bytegnn}
Chenguang Zheng, Hongzhi Chen, Yuxuan Cheng, Zhezheng Song, Yifan Wu, Changji
  Li, James Cheng, Hao Yang, and Shuai Zhang.
\newblock Bytegnn: efficient graph neural network training at large scale.
\newblock {\em Proceedings of the VLDB Endowment}, 15(6):1228--1242, 2022.

\bibitem{tgl}
Hongkuan Zhou, Da~Zheng, Israt Nisa, Vasileios Ioannidis, Xiang Song, and
  George Karypis.
\newblock Tgl: A general framework for temporal gnn training on billion-scale
  graphs.
\newblock {\em arXiv preprint arXiv:2203.14883}, 2022.

\bibitem{sgpstudy}
Zainab Abbas, Vasiliki Kalavri, Paris Carbone, and Vladimir Vlassov.
\newblock Streaming graph partitioning: an experimental study.
\newblock {\em Proceedings of the VLDB Endowment}, 11(11):1590--1603, 2018.

\end{thebibliography}
